\documentclass{article}
\usepackage{ijcai21}

\usepackage{times}
\usepackage{soul}
\usepackage{url}
\usepackage[hidelinks]{hyperref}
\usepackage[utf8]{inputenc}
\usepackage[small]{caption}
\usepackage{graphicx}
\usepackage{amsmath}
\usepackage{amsthm}
\usepackage{booktabs}
\usepackage{algorithm}
\usepackage{algorithmic}
\usepackage{algorithm,algorithmic,amsmath}
\usepackage{graphicx}
\usepackage{xcolor}
\usepackage{amssymb}
\usepackage{amsmath,bm}
\usepackage{algorithmic}
\usepackage{graphicx}
\usepackage{textcomp}
\usepackage{xcolor}
\usepackage{url}
\usepackage{wrapfig}
\usepackage{amsthm,multirow}
\newcommand{\bx}{\mathbf{x}}

\newcommand{\bw}{\mathbf{w}}

\newcommand{\CR}{\mathcal{R}}
\newcommand{\CB}{\mathcal{B}}

\newcommand{\CI}{\mathcal{I}}
\newcommand{\CL}{\mathcal{L}}

\newcommand{\BR}{\mathbb{R}}

\newcommand{\BE}{\mathbb{E}}
\newcommand{\BW}{\mathbf{W}}

\DeclareMathOperator*{\argmax}{\arg\!\max}
\DeclareMathOperator*{\argmin}{\arg\!\min}

\usepackage{hyperref}

\usepackage{amssymb}

\let\emptyset\varnothing
\usepackage{bbm}
\title{BOBCAT: Bilevel Optimization-Based Computerized Adaptive Testing}

\author{
    Aritra Ghosh and Andrew Lan
    \affiliations
    University of Massachusetts Amherst
    \emails
    \{arighosh,andrewlan\}@cs.umass.edu
}

\begin{document}

\maketitle

\begin{abstract}
Computerized adaptive testing (CAT) refers to a form of tests that are personalized to every student/test taker. CAT methods adaptively select the next most informative question/item for each student given their responses to previous questions, effectively reducing test length. Existing CAT methods use item response theory (IRT) models to relate student ability to their responses to questions and static question selection algorithms designed to reduce the ability estimation error as quickly as possible; therefore, these algorithms cannot improve by learning from large-scale student response data. In this paper, we propose BOBCAT, a Bilevel Optimization-Based framework for CAT to directly learn a data-driven question selection algorithm from training data. BOBCAT is agnostic to the underlying student response model and is computationally efficient during the adaptive testing process. Through extensive experiments on five real-world student response datasets, we show that BOBCAT outperforms existing CAT methods (sometimes significantly) at reducing test length. 
\end{abstract}

\section{Introduction}

One important feature of computerized/online learning platforms is computerized adaptive testing (CAT), which refers to tests that can accurately measure the ability/knowledge of a student/test taker using few questions/items, by using an algorithm to adaptively select the next question for each student given their response to previous questions \cite{cat,stevereport}. 
An accurate and efficient estimate of a student's knowledge levels helps computerized learning platforms to deliver personalized learning experiences for every learner. 

A CAT system generally consists of the following components: an underlying psychometric model that links the question's features and the student's features to their response to the question, a bank of questions with features learned from prior data, and an algorithm that selects the next question for each student from the question bank and decides when to stop the test; see \cite{health} for an overview. Most commonly used response models in CAT systems are item response theory (IRT) models, with their simplest form (1PL) given by
\begin{align} \label{eq:irt}
    p(Y_{i,j} = 1) = \sigma(\theta_i - b_j),
\end{align}
where $Y_{i,j}$ is student~$i$'s binary-valued response to question~$j$, where $1$ denotes a correct answer, $\sigma(\cdot)$ is the sigmoid/logistic function, and $\theta_i \in \mathbb{R}$ and $b_j \in \mathbb{R}$ are scalars corresponding to the student's ability and the question's difficulty, respectively \cite{lordirt,rasch}. More complex IRT models use additional question features such as the scale and guessing parameters or use multidimensional student features, i.e., their knowledge levels on multiple skills \cite{mirt}. 

Most commonly used question selection algorithms in CAT systems select the most informative question that minimizes the student feature measurement error; see \cite{information} for an overview. Specifically, in each step of the adaptive testing process (indexed by $t$) for student $i$, they select the next question as
\begin{align} \label{eq:selection}
    j_i^{(t)} = \textstyle \argmax_{j \in \Omega_i^{(t)}} I_j(\hat{\theta}_i^{(t-1)}),
\end{align}
where $\Omega_i^{(t)}$ is the set of available questions to select for this student at time step $t$ (the selected question at each time step is removed afterwards), $\hat{\theta}_i^{(t-1)}$ is the current estimate of their ability parameter given previous responses $Y_{i,j_i^{(1)}}, \ldots, Y_{i,j_i^{(t-1)}}$, and $I_j(\cdot)$ is the informativeness of question $j$. 
In the context of 1PL IRT models, most informativeness metrics will select the question with  difficulty closest to the current estimate of the student's ability, i.e., selecting the question that the student's probability of answering correctly is closest to $50\%$. This criterion coincides with uncertainty sampling \cite{us} for binary classification, a commonly used method in active learning \cite{alsurvey} that is deployed in real-world CAT systems \cite{duolingo-active}. 

Despite the effectiveness of existing CAT methods, two limitations hinder their further improvement. First, most question selection algorithms are specifically designed for IRT models (\ref{eq:irt}). The \emph{highly structured} nature of IRT models enables theoretical characterization of question informativeness but limits their ability to capture complex student-question interactions compared to more flexible, deep neural network-based models \cite{dirt,ncd}. This limitation is evident on large-scale student response datasets (often with millions of responses) that have been made available \cite{ednet,neuripschal}. Second, most existing question selection algorithms are \emph{static} since they require a predefined informativeness metric (\ref{eq:selection}); they can only use large-scale student response data to improve the underlying IRT model (e.g., calibrating question difficulty parameters) but not the question selection algorithm. 
Therefore, they will not significantly improve over time as more students take tests. Recently, there are ideas on using reinforcement learning to learn question selection algorithms \cite{rlcat,changdeep}; however, these methods have not been validated on real data.

\subsection{Contributions}
In this paper, we propose BOBCAT, a Bilevel Optimization-Based framework for Computerized Adaptive Testing. BOBCAT is based on the key observation that the ultimate goal of CAT is to reduce test length. Therefore, estimating student ability parameters is a \emph{proxy} of the real objective: \emph{predicting a student's responses to all questions on a long test that cannot be feasibly administered}. We make three key contributions: 

First, we recast CAT as a bilevel optimization problem \cite{bilevel} in the meta learning \cite{maml} setup: in the \emph{outer-level} optimization problem, we learn both the response model parameters and a data-driven question selection algorithm by \emph{explicitly} maximizing the predictive likelihood of student responses in a held-out \emph{meta} question set. In the \emph{inner-level} optimization problem, we adapt the outer-level response model to each student by maximizing the predicted likelihood of their responses in an observed \emph{training} question set. This bilevel optimization framework directly learns an effective and efficient question selection algorithm through the training-meta setup. Moreover, BOBCAT is agnostic to the underlying response model, compatible with both IRT models and deep neural network-based models; Once learned, the question selection algorithm selects the next question from past question responses directly, without requiring the student parameters to be repeatedly estimated in real time during the CAT process.

Second, we employ a \emph{biased} approximate estimator of the gradient w.r.t.\ the question selection algorithm parameters in the bilevel optimization problem. This approximation leverages the influence of each question on the algorithm parameters \cite{influence} to reduce the variance in the gradient estimate and leads to better question selection algorithms than an unbiased gradient estimator. 

Third, we verify the effectiveness of BOBCAT through extensive quantitative and qualitative experiments on five large-scale, real-world student response datasets. We observe that the learned data-driven question selection algorithms outperform existing CAT algorithms at reducing test length, requiring 50\% less questions to reach the same predictive accuracy on meta question set in some cases; this improvement is generally more significant on larger datasets. Our implementation will be publicly available at {\url{https://github.com/arghosh/BOBCAT}}.

\paragraph{Remarks.} We emphasize that the goal of this paper is to learn data-driven question selection algorithms, \textbf{not} to develop the best underlying student response model. We also acknowledge that BOBCAT may not be directly applicable to real-world CAT settings since its data-driven nature makes it hard to i) theoretically analyze and ii) prone to biases in historical data. Nevertheless, our promising results show that it is possible to improve the efficiency of CAT with large-scale student response data. Additionally, other important CAT aspects such as question exposure control \cite{exposure} need to be further investigated; we provide preliminary qualitative analyses in Section~\ref{sec:qualitative}.

\section{The BOBCAT Framework}
\label{sec:framework}
 \begin{figure}[tp]
    \centering
 \includegraphics[width=.48\textwidth]{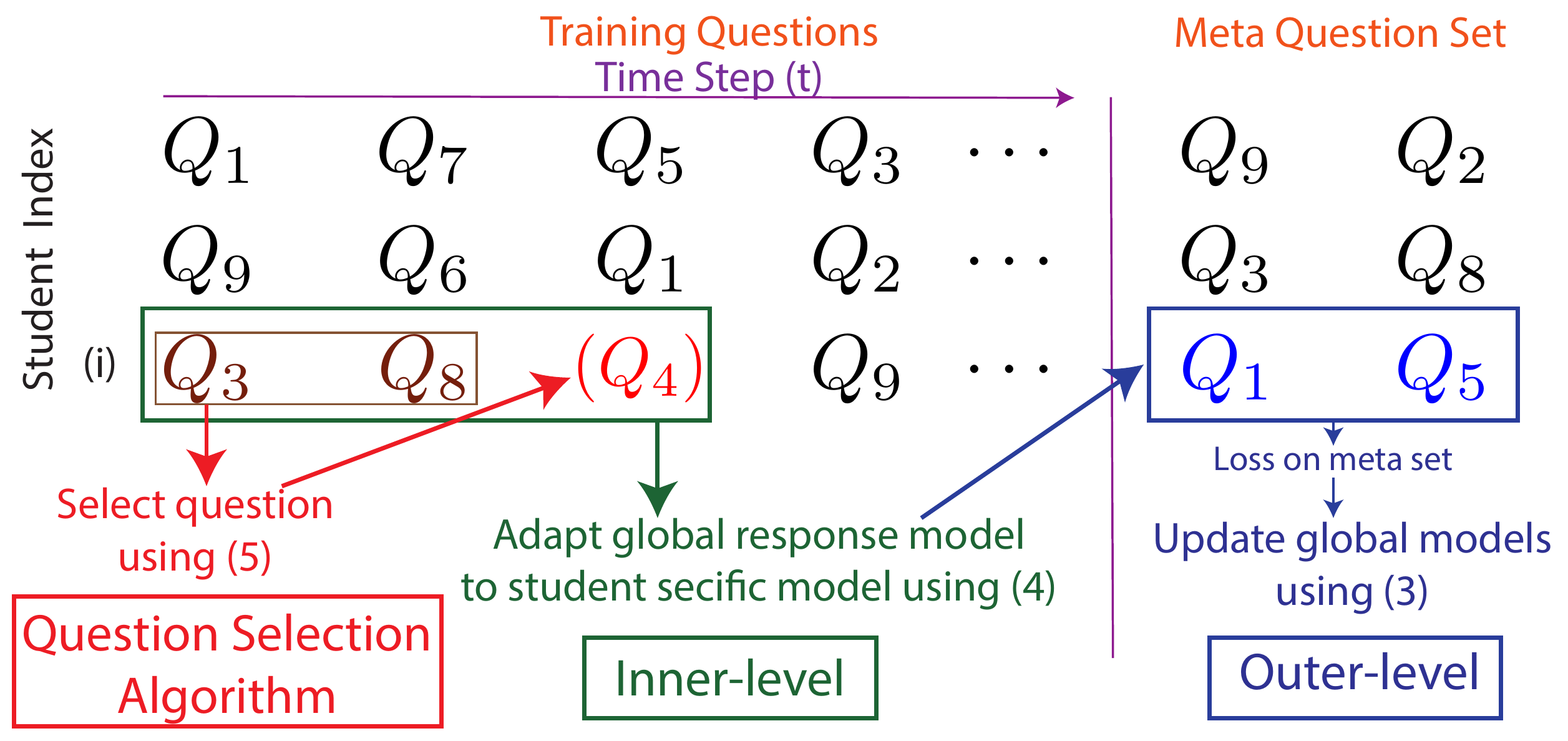}
    
\caption{Overview of the BOBCAT framework. 
}
    \label{fig:fw}
\end{figure}

We now detail the BOBCAT framework, visualized in Figure~\ref{fig:fw}. Let $N$ and $Q$ denote the number of students and questions in the student response dataset we use to train BOBCAT, respectively. For a single student $i$, we sequentially select a total of $n \, (\ll |\Omega_i^{(1)}|)$ questions\footnote{BOBCAT is applicable to both fixed-length and variable-length CAT designs \cite{stoprule}; we study the former in this paper.}, $\{j_i^{(1)},\cdots, j_i^{(n)}\}$, observe their responses, and predict their response on a held-out set of meta questions, $\Gamma_i$; $\Omega_i^{(1)}$ denotes the initial set of available questions and $\Omega_i^{(1)} \cap \Gamma_i=\emptyset$. 
The training and meta question sets are randomly selected and not the same for each student in the dataset. 
We solve the following bilevel optimization problem \cite{bilevel,implicit-gradient}:
\begin{align}
  & \underset{\bm{\gamma},\bm{\phi}}{\text{minimize}} \;\, \frac{1}{N}\!\sum_{i=1}^{N}\!\sum_{j\in \Gamma_i}\!\ell \Big(\!Y_{i,j}, g(j;\!{\bm{\theta}}_i^{\ast})\!\Big) \!\!:=\!\! \frac{1}{N}\!\sum_{i=1}^{N}\! \CL({\bm{\theta}}_i^{\ast},\!\Gamma_i) \label{eq:bilevel} \\
  & \mbox{s.t.} \;\, {\bm{\theta}}_i^{\ast}\!=\!\!\argmin_{\bm{\theta}_i} \! \sum_{t=1}^n \! \ell \Big(\!Y_{i,j_i^{(t)}}, g(j_i^{(t)};{\bm{\theta}_i})\!\Big)\!\!+\!\!\CR(\bm{\gamma}, {\bm{\theta}_i})\!\!:=\!\!\CL'(\bm{\theta}_i)\label{eq:inner} \\
  & \mbox{where} \;\, j_i^{(t)} \sim \Pi (Y_{i,j_i^{(1)}}, \ldots, Y_{i,j_i^{(t-1)}}; \bm{\phi}) \in \Omega_i^{(t)}. \label{eq:algo}
\end{align}
Here, $\bm{\gamma}$ and $\bm{\phi}$ are the \emph{global} response model and question selection algorithm parameters, respectively. $g(\cdot)$ is the response model, %
which takes as input the index of the question of interest, $j$, and 
uses the \emph{local} parameter specific to student $i$, $\bm{\theta}_i^*$, to output the prediction of the student's likelihood of responding to the question correctly. $\Pi(\cdot)$ is the question selection algorithm (red box in Figure~\ref{fig:fw}), which takes as input the student's responses to previously selected questions and outputs the index of the next selected question. 

The outer-level optimization problem (blue box in Figure~\ref{fig:fw}) minimizes the binary cross-entropy loss, $\ell(\cdot)$, on the \emph{meta} question sets across all students to learn both the global response model and the question selection algorithm; $\CL(\cdot)$ corresponds to the sum of this loss over questions each student responded to in the meta question set. The inner-level optimization problem (green box in Figure~\ref{fig:fw}) minimizes $\CL'(\cdot)$, the cross-entropy loss on a small number of questions selected for each student on the \emph{training} question set to adapt the global response model to each local student, resulting in a student-specific parameter $\bm{\theta}_i^*$; $\CR(\bm{\gamma},\bm{\theta}_i)$ is a regularization term that penalizes large deviations of the local parameters from their global values. Note that $\bm{\theta}_i^*$ is a function of the global parameters $\bm{\gamma}$ and $\bm{\phi}$, reflected through both the regularization term in (\ref{eq:inner}) and the question selection algorithm through questions it selects for this student in (\ref{eq:algo}).

\paragraph{Response Model.} The response model $g(\cdot)$ can be taken as either IRT models or neural network-based models. In the case of IRT models, the global parameters $\bm{\gamma}$ corresponds to the combination of the question difficulties and the student ability prior. We adapt these parameters to each local student through their responses to selected questions in the inner-level optimization problem. In our experiments, we only use the global student ability as the prior mean of each local student's ability estimate and keep the question difficulties fixed in the inner-level optimization problem, following the typical setup in real-world CAT systems. In the case of neural network-based models, the parameters are usually not associated with any specific meaning; following standard practices in meta-learning \cite{meta-opt-fix}, we fix part of the network (e.g., all weights and biases, which one can regard as a nonlinear version of question difficulties) and optimize the rest of the network (e.g., the input vector, which one can regard as student abilities) in the inner-level optimization problem. 

\paragraph{Question Selection Algorithm.}
The question selection algorithm $\Pi(\cdot)$ can be either deterministic or probabilistic, i.e., it either outputs a single selected question or a probability distribution over available questions. 
We define the input \emph{state} vector to the question selection algorithm at step $t$ as $\bx_i^{(t)} \in \{-1,0,1\}^Q$, where an entry of $-1$ denotes an incorrect response to a past selected question, $1$ denotes a correct response, while $0$ denotes questions that have not been selected. 
We do not include the time step at which a question is selected in the state vector since in CAT settings, the student's true ability is assumed to be static during the testing process while an estimate is being updated. %
Although any differentiable model architecture can be used for the question selection algorithm, we use the multi-layer perceptron model that is invariant to question ordering. 
For probabilistic question selection algorithms, we select a question by sampling from the output distribution $j^{(t)}_i \sim \Pi(\bx^{(t)}_i, \Omega_i^{(t)};\bm{\phi})$.

\subsection{Optimization}
\label{sec:policy}
We use gradient descent (GD) to solve the inner-level optimization problem for the local response model parameters $\bm{\theta}_i^{\ast}$, following model-agnostic meta learning \cite{maml}.  In particular, we let the local student-specific parameter deviate from the global response model parameters  %
by taking $K$ GD steps from $\bm{\gamma}$, where each step is given as
\begin{align}
\bm{\theta}_i\leftarrow {\bm{\theta}}_i-\alpha\nabla_{{\bm{\theta}}}\sum_{t=1}^n\ell\Big(Y_{i,j_i^{(t)}}, {g} (j_i^{(t)};\bm{\theta})   \Big)\Big|_{{\bm{\theta}}_i},\label{eq:inner-maml}
\end{align}
where $\alpha$ is the learning rate. We do not explicitly use regularization for GD steps since early stopping (with only a few GD steps) is equivalent to a form of regularization \cite{implicit-gradient}. 
Computing the gradient w.r.t.\ the global parameters $\bm{\gamma}$ requires us to compute the gradient w.r.t.\ the gradient in the inner-level optimization problem in (\ref{eq:inner-maml}) (also referred to as the meta-gradient), which can be computed using automatic differentiation \cite{automatic}. Computing the exact meta-gradient requires second-order derivatives; however, we found that first-order approximation works well in practice and leads to low computational complexity. 
Similarly, to learn the selection algorithm parameters $\bm{\phi}$, we need to compute the gradient of the outer-level objective in (\ref{eq:bilevel}) %
w.r.t.\ $\bm{\phi}$ through the student-specific parameters $\bm{\theta}^{\ast}_i(\bm{\gamma}, \bm{\phi})$, i.e., the solution to the inner-level optimization problem. The gradient for a single student $i$ (the full gradient sums across all students) is given by 
\begin{align}
\resizebox{0.91\hsize}{!}{%
 $\nabla_{\bm{\phi}}\CL\Big({\bm{\theta}}_i^{\ast} \big(\bm{\gamma}, \bm{\phi}\big), \Gamma_i  \Big)\!=\!
 \nabla_{\bm{\phi}}\BE_{j_i^{(1:n)}\sim \Pi(\cdot; {\bm{\phi}})} \Big[\CL\Big({\bm{\theta}}_i^{\ast} \big(\bm{\gamma}, \{j_i^{(1:n)}\} \big), \Gamma_i  \Big)\Big]$\label{eq:phi-grad},
}
\end{align}
where we replace the dependence of $\bm{\theta}^{\ast}_i$ on the parameters of the question selection algorithm, $\bm{\phi}$, with the indices of the selected questions, $j_i^{(1:n)}$, which we need to backpropagate through. The discrete nature of these variables makes them non-differentiable so that we cannot compute the exact gradient. Next, we will detail two ways to estimate this gradient.

\subsubsection{Unbiased Gradient Estimate}
We can use the score function-based identity ($\frac{\partial \log f(X;\bm{\phi})}{\partial \bm{\phi}} = \frac{{\partial f(X;\bm{\phi})}/{\partial \bm{\phi}}}{f(X;\bm{\phi})} $ for any probability distribution $f(X;\bm{\phi})$) to estimate the unbiased gradient in (\ref{eq:phi-grad}) \cite{reinforce} as

\begin{align}
    &\nabla_{\bm{\phi}}\BE_{j_i^{(1:n)}\sim \Pi(\cdot;{\phi})} \Big[\CL\Big({\bm{\theta}}_i^{\ast} \big(\bm{\gamma},  \{j_i^{(1:n)}\}\big), \Gamma_i  \Big)\Big]\label{eq:reinforce}\\
=&\BE_{j_i^{(1:n)}\sim \Pi(\cdot;{\phi})}\! \Big[\big(\CL({\bm{\theta}}_i^{\ast},\Gamma_i) -b_i \big)\nabla_{\bm{\phi}}\!\log\! \prod_{t=1}^n \Pi (j_i^{(t)}|\bx^{(t)}_i;\!{\bm{\phi}})\!\Big]\nonumber\\
=&\BE_{j_i^{(1:n)}\sim \Pi(\cdot;{\phi})}\! \Big[\big(\CL({\bm{\theta}}_i^{\ast},\Gamma_i) -b_i \big) \sum_{t=1}^n\! \nabla_{\bm{\phi}}\!\log \! \Pi (j_i^{(t)}|\bx^{(t)}_i;\!{\bm{\phi}})\!\Big] ,\nonumber
\end{align}
where $b_i$ is a control variable to the reduce the variance of the gradient estimate. 
This unbiased gradient resembles reinforcement learning-type algorithms for CAT, an idea discussed in \cite{rlcat}. 
We use proximal policy optimization for its training stability with an actor network and a critic network \cite{ppo}; we provide details of these networks in the supplementary material to be released in the full version of the paper.

We observe that this unbiased gradient estimate updates the question selection algorithm parameters through the selected questions only, without including observations on the available but not selected questions, resulting in slow empirical convergence in practice. 
However, incorporating information on unselected questions into the gradient computation may lead to lower variance in the gradient and stabilize the training process. Next, we detail a biased approximation to the gradient using all the available training questions.

\subsubsection{Approximate Gradient Estimate}
We can rewrite the gradient in (\ref{eq:phi-grad}) as
\begin{align}
\nabla_{\bm{\phi}}\CL\Big({\bm{\theta}}_i^{\ast} \big(\bm{\gamma}, \bm{\phi}\big), \Gamma_i  \Big)
=\nabla_{\bm{\theta}_i^{\ast}} \CL\Big({\bm{\theta}}_i^{\ast}, \Gamma_i  \Big)
\nabla_{\bm{\phi}} {\bm{\theta}}_i^{\ast} \big(\bm{\gamma}, \bm{\phi}\big).\label{eq:split-grad}
\end{align}
The gradient w.r.t.\ ${\bm{\theta}_i^{\ast}}$ can be computed exactly; next, we discuss the computation of $\nabla_{\bm{\phi}} {\bm{\theta}}_i^{\ast} \big(\bm{\gamma}, \bm{\phi}\big)$ in detail for a single time step $t$. 
We can rewrite the inner-level optimization in (\ref{eq:inner}) by splitting the current question index $j_i^{(t)}$ from previously selected question indices $j_i^{(1)},\cdots, j_i^{(t-1)}$ as
\begin{align}
{\bm{\theta}}_i^{\ast}\!=\!\argmin_{\bm{\theta}_i} \! \sum_{\tau=1}^{t-1} \! \ell \Big(Y_{i,j_i^{(\tau)}}, g(j_i^{(\tau)};{\bm{\theta}_i})\Big)\!+\!\CR(\bm{\gamma}, {\bm{\theta}_i})\nonumber\\
+\sum_{j\in \Omega_i^{(t)}} w_j(\bm{\phi}) \ell \Big(Y_{i,j}, g(j;{\bm{\theta}_i})\Big) %
,\label{eq:inner-weighted}
\end{align}
where $w_j(\bm{\phi})=1$ if $j=j_i^{(t)}$ and $w_j(\bm{\phi})=0$ for all other available questions. %
In (\ref{eq:inner-weighted}), we can compute the derivative $\frac{d \bm{\theta}_i^{\ast}}{d  w_j(\bm{\phi})}$ for all available question indices in $\Omega_i^{(t)}$ regardless of whether they are selected at time step $t$, using the implicit function theorem \cite{cook1982residuals} as
\begin{align*}
\frac{d \bm{\theta}_i^{\ast}}{d w_j(\bm{\phi})} = - 
\Big(\nabla_{\bm{\theta}_i}^2 \CL'_i\Big)^{-1}
 \nabla_{\bm{\theta}_i} \ell \Big(Y_{i,j}, g(j;{\bm{\theta}_i})\Big)
\Big|_{{\bm{\theta}}_i^{\ast}}.
\end{align*}
This gradient can be computed without explicitly computing the inverse Hessian matrix using
automatic differentiation in a way similar to that for the global response model parameters $\bm{\gamma}$.
However, we still need to compute $\frac{\partial w_j(\bm{\phi})}{\partial \Pi(j|\bx_i^{(t)};\bm{\phi})}$, which is not differentiable; since $w_j(\bm{\phi})= \Pi(j|\bx_i^{(t)};\bm{\phi})$ holds when the selection algorithm network puts all the probability mass on a single question, we can use the approximation  $w_j(\bm{\phi})\approx \Pi(j|\bx_i^{(t)};\bm{\phi})$. From (\ref{eq:split-grad}) and (\ref{eq:inner-weighted}), it turns out that under this approximation, the full gradient with respect to a single question, %
$\frac{\partial \CL(\bm{\theta}_i^{\ast}, \Gamma_i)}{\partial \Pi(j|\bx_i^{(t)};\bm{\phi})}$, is the widely used influence function score \cite{influence}:
\begin{align}
 -\!\nabla_{\bm{\theta}_i}\!\CL(\bm{\theta}_i,\! \Gamma_i\!)\!
 \Big(\!\nabla_{\bm{\theta_i}}^2\! \CL'_i\Big)^{\!-\!1}\!
 \!\nabla_{\bm{\theta}_i}\! \ell \Big(\!Y_{i,j}, g(\!j;\!{\bm{\theta}_i}\!)\!\Big)\!
\Big|_{{\bm{\theta}}_i^{\ast}}\!\! :=\! \CI_i(j),\label{eq:approx}
\end{align}
where $\CI_i(j)$, the influence function score of question $j$, computes the change in the loss on the meta question set under small perturbations in the weight of this question, $w_j(\bm{\phi})$, in (\ref{eq:inner-weighted}). 
Intuitively, we would want to select available training questions with gradients that are similar to the gradient on the meta question set, i.e., those with the most information on meta questions; the approximation enables us to learn such a question selection algorithm by backpropagating the influence score as gradients through all available questions in the training question set. 
In contrast, for the unbiased gradient in (\ref{eq:reinforce}), $\frac{\partial \CL(\bm{\theta}_i^{\ast}, \Gamma_i)}{\partial \Pi(j|\bx_i^{(t)};\bm{\phi})}$ equals zero for all unselected questions and equals 
$  -(\CL(\bm{\theta}_i^{\ast},\Gamma_i)-b_i) \log \Pi(j_i^{(t)}|\bx_i^{(t)};\bm{\phi})$
for the selected question $j_i^{(t)}$. 
This biased approximation (often known as the straight-through estimator) has been successfully applied in previous research for neural network quantization and leads to lower empirical variance \cite{st}. 
Algorithm~\ref{alg:training} summarizes BOBCAT's training process. 

\begin{algorithm}[t]
\begin{algorithmic}[1]
\STATE Initialize global parameters $\bm{\gamma}, \bm{\phi}$, learning rates $\eta_1,\eta_2, \alpha$, and number of GD steps at the inner-level, $K$.
\WHILE {not converged} 
\STATE Randomly sample a mini-batch of students $\CB$ with training and meta question sets $\{\Omega_i^{(1)},\Gamma_i\}_{i\in\CB}$.
\FOR{$t\in 1\dots n$}
\STATE Encode the student's current state $\bx^{(t)}_i$ based on their responses to previously selected questions.
\STATE Select question $j_i^{(t)}\sim\Pi(\bx^{(t)}_i;\bm{\phi})$ for each student.
\STATE Optimize ${\bm{\theta}}^{\ast}_i$ in Eq.~\ref{eq:inner-maml} using learning rate $\alpha$ and $K$ GD steps on observed responses $\{Y_{i,j_i^{(1:t)}}\}$.
\STATE Estimate the unbiased (or the approximate) gradient $\nabla_{\bm{\phi}} \CL(\bm{\theta}_i^{\ast}, \Gamma_i)$ using Eq.~\ref{eq:reinforce} (or Eq.~\ref{eq:approx}).
\STATE Update $\bm{\phi}$:  $\bm{\phi}\leftarrow \bm{\phi}-\frac{\eta_2}{|\CB|}\sum_{i\in \CB}\nabla_{\bm{\phi}}\CL(\bm{\theta}_i^{\ast}, \Gamma_i)$.
\ENDFOR
\STATE Update $\bm{\gamma}$:  $\bm{\gamma}\!\leftarrow\!\bm{\gamma}\!-\!\frac{\eta_1}{|\CB|}\!\sum_{i\in \CB}\!\nabla_{\bm{\gamma}}\CL\Big(\bm{\theta}_i^{\ast}(\bm{\gamma},\bm{\phi}), \Gamma_i\!\Big)$.
\ENDWHILE 
\end{algorithmic}
\caption{BOBCAT training process}
\label{alg:training}
\end{algorithm}

\paragraph{Computational Complexity.} 
At training time, we need to solve the full BOBCAT bilevel optimization problem, which is computationally intensive on large datasets. 
However, at test time, when we need to select the next question for each student, we only need to use their past responses as input to the learned question selection algorithm $\Pi(\cdot;{\bm{\phi}})$ to get the selected question as output; this operation is more computationally efficient than existing CAT methods that require updates to the student's ability estimate after every question.

\section{Experiments}
\label{sec:exp}

We now detail both quantitative and qualitative experiments we conducted on five real-world student response datasets to validate BOBCAT's effectiveness. 

\paragraph{Datasets, Training, Testing and Evaluation Metric.} 
We use five publicly available benchmark datasets: EdNet\footnote{ \href{https://github.com/riiid/ednet}{https://github.com/riiid/ednet}}, 
Junyi\footnote{\href{https://www.kaggle.com/junyiacademy/learning-activity-public-dataset-by-junyi-academy}{https://www.kaggle.com/junyiacademy/learning-activity-public-dataset-by-junyi-academy}},
Eedi-1, Eedi-2\footnote{\href{https://eedi.com/projects/neurips-education-challenge}{https://eedi.com/projects/neurips-education-challenge}},
and ASSISTments\footnote{\href{https://sites.google.com/site/assistmentsdata/home/assistment-2009-2010-data}{https://sites.google.com/site/assistmentsdata/home/assistment-2009-2010-data}}.
In Table~\ref{tab:dataset}, we list the number of students, the number of questions, and the number of interactions.
We provide preprocessing details and additional background on each dataset in the supplementary material.
We perform 5-fold cross validation for all datasets; for each fold, we use 60\%-20\%-20\% \emph{students} for training, validation, and testing, respectively. For each fold, we use the validation students to perform early stopping %
and tune the parameters for every method. 
For BOBCAT, we partition the questions responded to by each student into the training ($\Omega_i^{(1)}$, 80\%) and meta ($\Gamma_i$, 20\%) question sets. 
To prevent overfitting, we randomly generate these partitions in each training epoch.
We use both accuracy and the area under the receiver operating characteristics curve (AUC) as metrics to evaluate the performance of all methods on predicting binary-valued student responses on the meta set $\Gamma_i$. 
We implement all methods in $\texttt{PyTorch}$ and run our experiments in a $\texttt{NVIDIA TitanX/1080Ti}$ GPU.%

  \begin{table}[t]

  \centering
 \scalebox{0.77}{
     \begin{tabular}{p{0.7cm}c|ccccc}\toprule
{\scriptsize Dataset} & { n} & {\scriptsize IRT-Active} & {\scriptsize BiIRT-Active} & {\scriptsize BiIRT-Unbiased} & {\scriptsize BiIRT-Approx} & {\scriptsize BiNN-Approx} \\ \toprule
\multirow{4}{*}{EdNet} & 1  & 70.08 & 70.92 & 71.12 & \bf{71.22 }  & \bf{71.22} \\
 & 3  & 70.63 & 71.16 & 71.3 & 71.72 & \bf{71.82 }  \\
 & 5  & 71.03 & 71.37 & 71.45 & 71.95 & \bf{72.17 }  \\
 & 10  & 71.62 & 71.75 & 71.79 & 72.33 & \bf{72.55 }  \\
\midrule
\multirow{4}{*}{Junyi} & 1  & 74.52 & 74.93 & 74.97 & \bf{75.11 }  & 75.1 \\
 & 3  & 75.19 & 75.48 & 75.53 & 75.76 & \bf{75.83 }  \\
 & 5  & 75.64 & 75.79 & 75.75 & 76.11 & \bf{76.19 }  \\
 & 10  & 76.27 & 76.28 & 76.19 & 76.49 & \bf{76.62 }  \\
\midrule
\multirow{4}{*}{Eedi-1} & 1  & 66.92 & 68.22 & 68.61 & \bf{68.82 }  & 68.78 \\
 & 3  & 68.79 & 69.45 & 69.81 & 70.3 & \bf{70.45 }  \\
 & 5  & 70.15 & 70.28 & 70.47 & 70.93 & \bf{71.37 }  \\
 & 10  & 71.72 & 71.45 & 71.57 & 72.0 & \bf{72.33 }  \\
\midrule
\multirow{4}{*}{Eedi-2} & 1  & 63.75 & 64.83 & 65.22 & 65.3 & \bf{65.65 }  \\
 & 3  & 65.25 & 66.42 & 67.09 & 67.23 & \bf{67.79 }  \\
 & 5  & 66.41 & 67.35 & 67.91 & 68.23 & \bf{68.82 }  \\
 & 10  & 68.04 & 68.99 & 68.84 & 69.47 & \bf{70.04 }  \\
\midrule
\multirow{4}{*}{\parbox{0.6cm}{ASSIST ments}} & 1  & 66.19 & 68.69 & 69.03 & \bf{69.17 }  & 68.0 \\
 & 3  & 68.75 & 69.54 & 69.78 & \bf{70.21 }  & 68.73 \\
 & 5  & 69.87 & 69.79 & 70.3 & \bf{70.41 }  & 69.03 \\
 & 10  & 71.04 & 70.66 & \bf{71.17 }  & 71.14 & 69.75 \\
\midrule
     \end{tabular}
}
\caption{Average predictive accuracy on the meta question set across folds on all datasets. Best methods are shown in \textbf{bold} font. For standard deviations and results on all methods, refer to  Figure~\ref{fig:acc} and Tables in the supplementary material. 
}
\label{tab:quant}
\end{table}

  \begin{table}[t]\centering
 \scalebox{0.77}{
     \begin{tabular}{p{0.7cm}c|ccccc}\toprule
{\scriptsize Dataset} & { n} & {\scriptsize IRT-Active} & {\scriptsize BiIRT-Active} & {\scriptsize BiIRT-Unbiased} & {\scriptsize BiIRT-Approx} & {\scriptsize BiNN-Approx} \\ \toprule
\multirow{4}{*}{EdNet} & 1  & 73.58 & 73.82 & 74.14 & 74.34 & \bf{74.41 }  \\
 & 3  & 74.14 & 74.21 & 74.49 & 75.26 & \bf{75.43 }  \\
 & 5  & 74.6 & 74.56 & 74.77 & 75.68 & \bf{76.07 }  \\
 & 10  & 75.35 & 75.21 & 75.39 & 76.35 & \bf{76.74 }  \\
\midrule
\multirow{4}{*}{Junyi} & 1  & 74.92 & 75.53 & 75.67 & \bf{75.91 }  & 75.9 \\
 & 3  & 76.06 & 76.52 & 76.71 & 77.11 & \bf{77.16 }  \\
 & 5  & 76.82 & 77.07 & 77.07 & 77.69 & \bf{77.8 }  \\
 & 10  & 77.95 & 77.95 & 77.86 & 78.45 & \bf{78.6 }  \\
\midrule
\multirow{4}{*}{Eedi-1} & 1  & 68.02 & 70.22 & 70.95 & \bf{71.34 }  & 71.33 \\
 & 3  & 71.63 & 72.47 & 73.26 & 74.21 & \bf{74.44 }  \\
 & 5  & 73.69 & 73.97 & 74.54 & 75.47 & \bf{76.0 }  \\
 & 10  & 76.12 & 75.9 & 76.34 & 77.07 & \bf{77.51 }  \\
\midrule
\multirow{4}{*}{Eedi-2} & 1  & 69.0 & 70.15 & 70.64 & 70.81 & \bf{71.24 }  \\
 & 3  & 71.11 & 72.18 & 73.11 & 73.37 & \bf{73.88 }  \\
 & 5  & 72.42 & 73.21 & 74.19 & 74.55 & \bf{75.2 }  \\
 & 10  & 74.36 & 75.17 & 75.37 & 75.96 & \bf{76.63 }  \\
\midrule
\multirow{4}{*}{\parbox{0.6cm}{ASSIST ments}} & 1  & 69.14 & 70.55 & 71.0 & \bf{71.33 }  & 70.12 \\
 & 3  & 71.17 & 71.6 & 72.35 & \bf{73.16 }  & 71.57 \\
 & 5  & 72.26 & 71.65 & 73.1 & \bf{73.71 }  & 72.14 \\
 & 10  & 73.62 & 72.52 & 74.38 & \bf{74.66 }  & 73.59 \\
\midrule
     \end{tabular}
}
\caption{Average AUC on the meta question set across folds on all datasets. For standard deviations and results on all methods, refer to Figures and Tables in the supplementary material.}
\label{tab:quant-auc}
\end{table}

\paragraph{Methods and Baselines.}
 \begin{figure}[!htb]
    \centering
 \includegraphics[width=1\columnwidth]{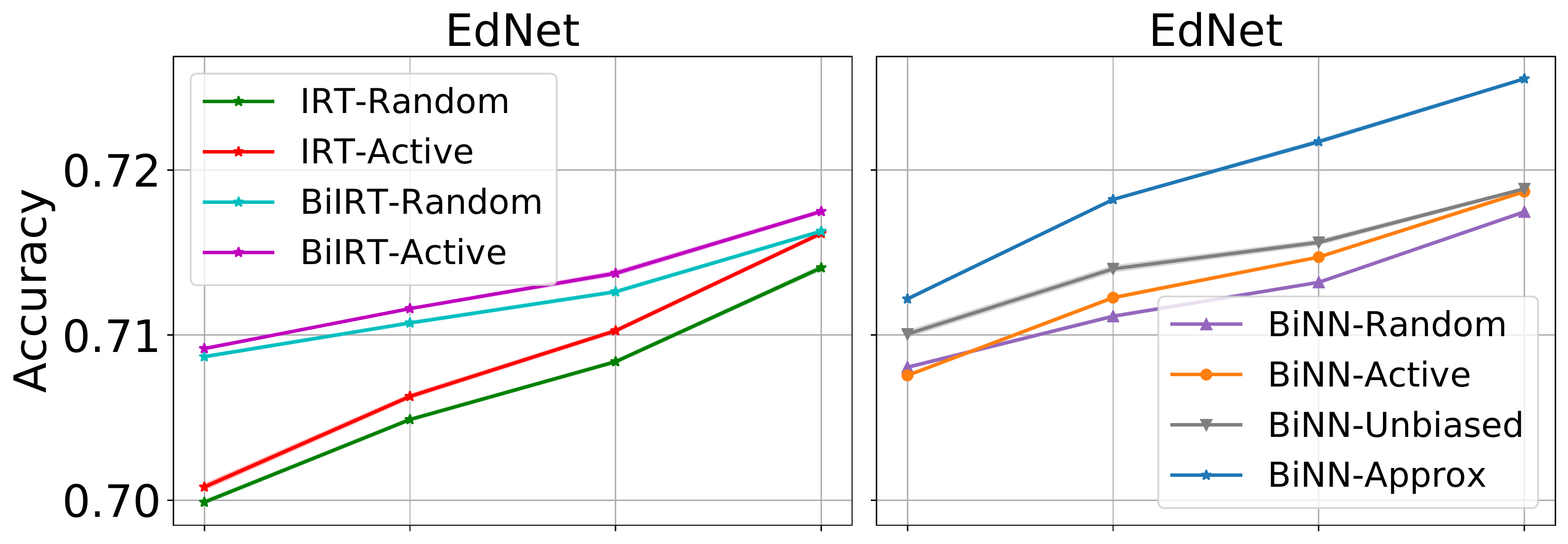}\\
    \includegraphics[width=\columnwidth]{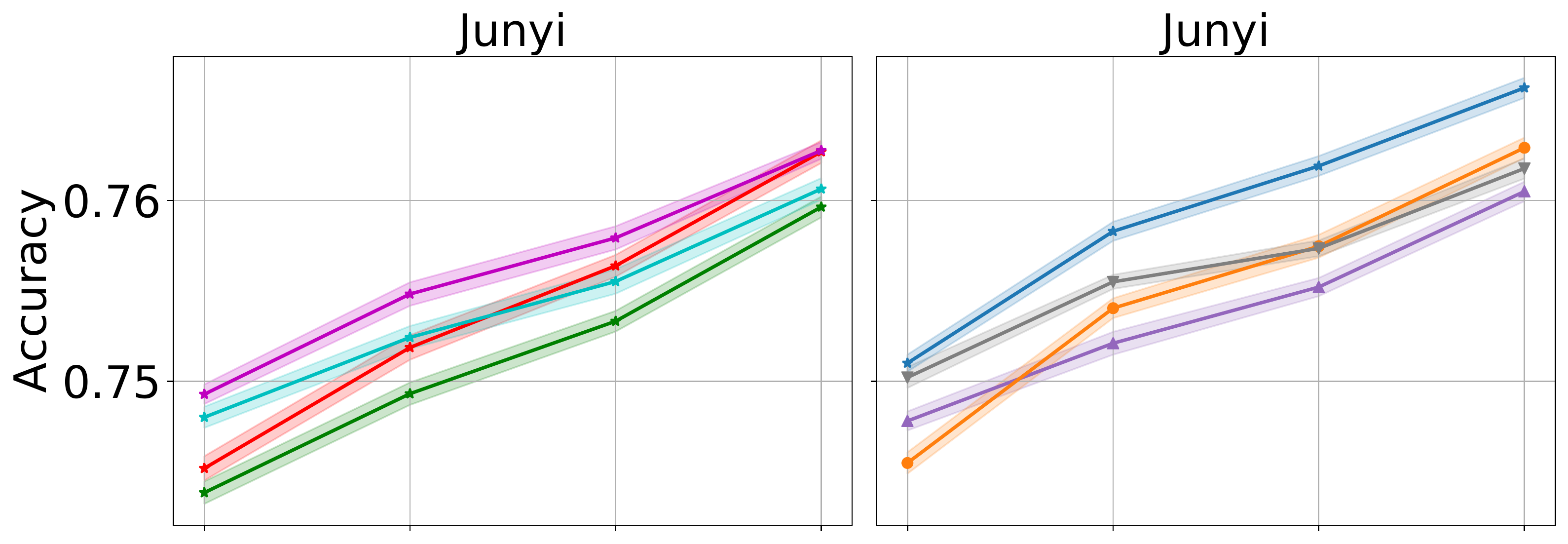}\\
    \includegraphics[width=\columnwidth]{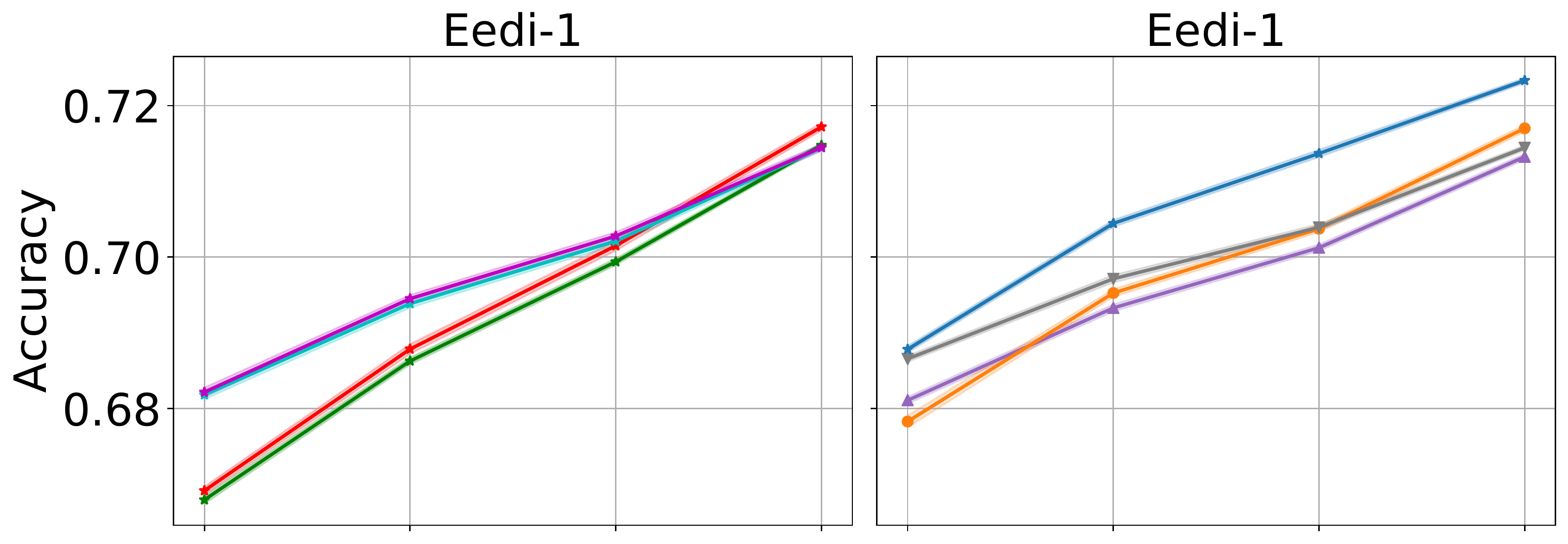}\\
    \includegraphics[width=\columnwidth]{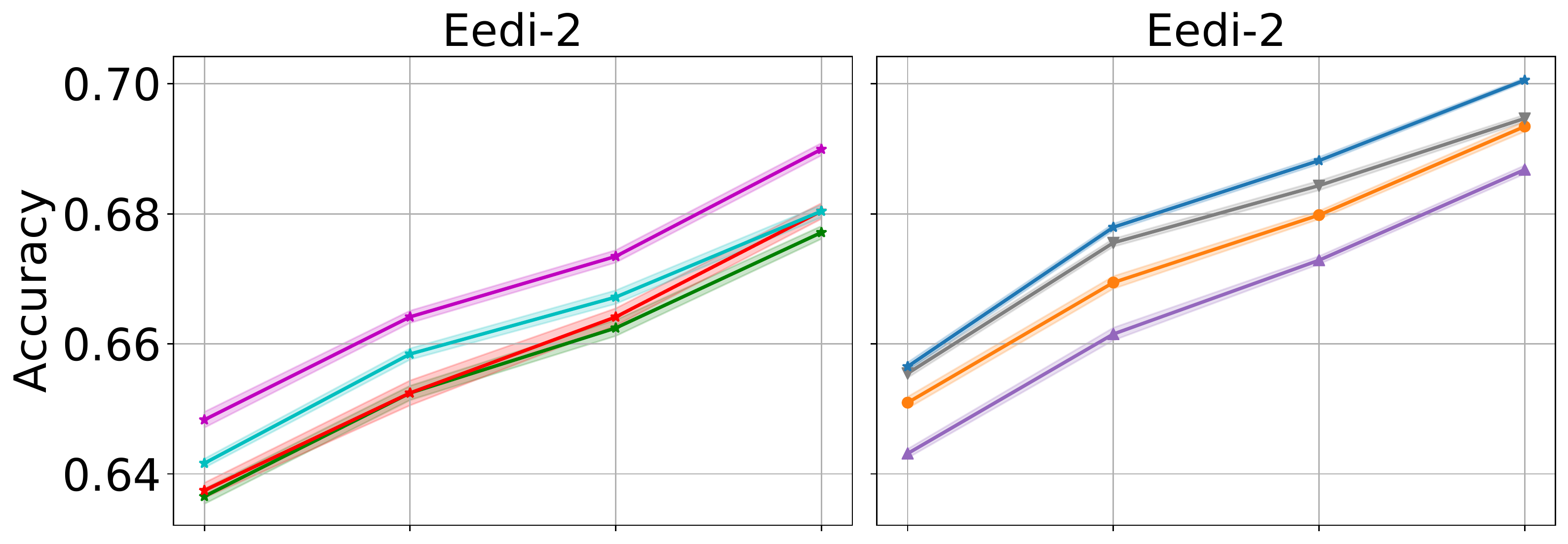}\\
    \includegraphics[width=\columnwidth]{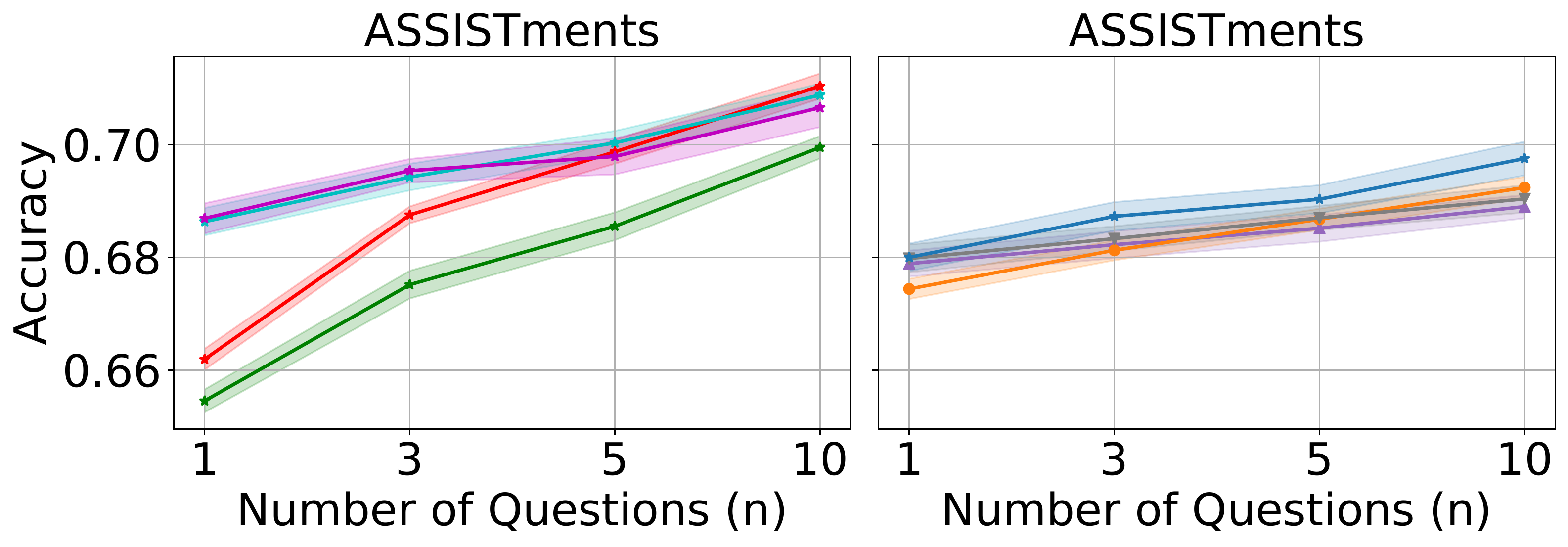}\\
    \caption{Average accuracy (dark lines) and 5-fold standard deviation (light fill lines) on all datasets. First column compares IRT vs BiIRT models; second column compares all BiNN models. 
    }
    \label{fig:acc}
\end{figure}

 \begin{table}[t]\centering
 \scalebox{0.88}{
     \begin{tabular}{c|ccccc}\toprule
    {\footnotesize \bf Dataset} &{\footnotesize  \bf EdNet} &{\footnotesize \bf Junyi} &{\footnotesize  \bf Eedi-1} &  {\footnotesize \bf Eedi-2}& {\footnotesize \bf ASSISTments} \\
    \toprule
    {\footnotesize \bf Students} & 
    312K & 52K & 119K & 5K &  2.3K \\
    {\footnotesize \bf Questions} & 
    13K & 25.8K & 27.6K & 1K & 26.7K\\
    {\footnotesize \bf Interactions} &
    76M & 13M & 15M &  1.4M & 325K\\
    \midrule
\end{tabular}
}
\caption{Dataset statistics. %
}
\label{tab:dataset}
\end{table}

For existing CAT methods, we use \textbf{IRT-Active}, the uncertainty sampling-based \cite{us} active learning question selection algorithm, which selects the next question with difficulty closest to a student's current ability estimate, as a baseline \cite{duolingo-active}. This method coincides with the question information-based CAT methods under the 1PL IRT model. 
We also use an additional baseline that selects the next question randomly, which we dub \textbf{IRT-Random}. 
For BOBCAT, we consider the cases of using IRT models (which we dub as \textbf{BiIRT}) and neural networks (which we dub as \textbf{BiNN}) as the response model. 
For both BiIRT and BiNN, we use four question selection algorithms: in addition to the \textbf{-Active} and \textbf{-Random} algorithms above, we also use learned  algorithms with the \textbf{-Unbiased} gradient (\ref{eq:reinforce}) and the approximate (\textbf{-Approx}) gradient (\ref{eq:approx}) on the question selection algorithm parameters $\bm{\phi}$.

\paragraph{Networks and Hyper-parameters.} 
We train IRT models using logistic regression with $l_2$-norm regularization. For IRT-Active, we compute the student's current ability estimate with $l_2$-norm regularization to penalize deviation from the mean student ability parameter. 
For BiNN, we use a two-layer,  fully-connected network (with 256 hidden nodes, ReLU nonlinearity, 20\% dropout rate, and a final sigmoid output layer) \cite{dlbook} as the response model, with a student-specific, 256-dimensional ability vector as input. 
We use another fully-connected network (with two hidden layers, 256 hidden nodes, Tanh nonlinearity, and a final softmax output layer) \cite{dlbook} as the question selection algorithm. 
For BiNN/IRT-Unbiased, we use another fully-connected critic network (two hidden layers, 256 hidden nodes, Tanh nonlinearity) in addition to the question selection actor network. 
For BiIRT and BiNN, we learn the global response model parameters $\bm{\gamma}$ and question selection algorithm parameters $\bm{\phi}$ 
using the Adam optimizer \cite{adam}
and learn the response parameters adapted to each student (in the inner-level optimization problem) using the SGD optimizer \cite{dlbook}. 
We provide specific hyper-parameter choices and batch sizes for each dataset in the supplementary material.
For all methods, we select $n\in \{1,3,5,10\}$ questions for each student.  %

\subsection{Results and Discussion}

In Table~\ref{tab:quant}, we list the mean accuracy numbers across all folds for selected BOBCAT variants and IRT-Active on all datasets; in Table~\ref{tab:quant-auc}, we do the same using the AUC metric. In the supplementary material, we provide results for all methods mentioned above and also list the standard deviations across folds. Using a neural network-based response model, BiNN-Approx outperforms other methods in most cases. Using an IRT response model, BiIRT-Approx performs similarly to BiNN-Approx and outperforms other methods. All BOBCAT variants significantly outperform IRT-Active, which uses a static question selection algorithm. On the ASSISTments dataset, the smallest of the five, BiIRT-Approx outperforms BiNN-Approx, which overfits. 
These results show that i) BOBCAT improves existing CAT methods by explicitly learning a question selection algorithm from data, where the improvement is more obvious on larger datasets, and ii) since BOBCAT is agnostic to the underlying response model, one can freely choose either IRT models when training data is limited or neural network-based models when there is plenty of training data.

In Figure~\ref{fig:acc}, we use a series of plots as ablation studies to present a more detailed comparison between different methods; here, we include random question selection as a bottom line. 
In the first column, we plot the mean and the standard deviation of accuracy for IRT-Random, IRT-Active, BiIRT-Random, and BiIRT-Active versus the number of questions selected. 
On the Eedi-1 dataset, BiIRT-Active performs better than IRT-Active on smaller $n$ but performs slightly worse for $n=10$. 
On the ASSISTments dataset, we observe a high standard deviation for larger $n$; nevertheless, BiIRT variants outperform IRT counterparts.  
On all other datasets, the BiIRT methods outperform their IRT counterparts significantly.
To reach the same accuracy, on the EdNet, Eedi-2, and Junyi datasets, BiIRT-Active requires $\!\sim\!30\%$ less questions compared to IRT-Active. 
This head-to-head comparison using IRT as the underlying response model demonstrates the power of bilevel optimization; even using static question selection algorithms, explicitly maximizing the predictive accuracy on a meta question set results in better performance, although the performance gain may not be significant. 

In the second column, we compare different BOBCAT variants using the same underlying neural network-based response model. 
We observe that on all datasets, BiNN-Approx significantly outperforms other methods, reaching the same accuracy as BiNN-Active with $50\%$-$75\%$ less questions. This performance gain is more significant on larger datasets. 
It also significantly outperforms the unbiased gradient estimate, reaching the same accuracy with $10\%$-$70\%$ less questions. 
BiNN-Unbiased significantly outperforms BiNN-Active for smaller $n$ but not for large $n$; we believe the large variance of the unbiased gradient might be the reason for this behavior. 
This head-to-head comparison shows that our approximate gradient estimate stabilizes the model training process and leads to better model fit. Moreover, data-driven question selection algorithms learned through bilevel optimization are much better than standard static CAT question selection algorithms and get better with more training data.

\begin{figure}[tp]
    \centering
    \includegraphics[width=.48\textwidth]{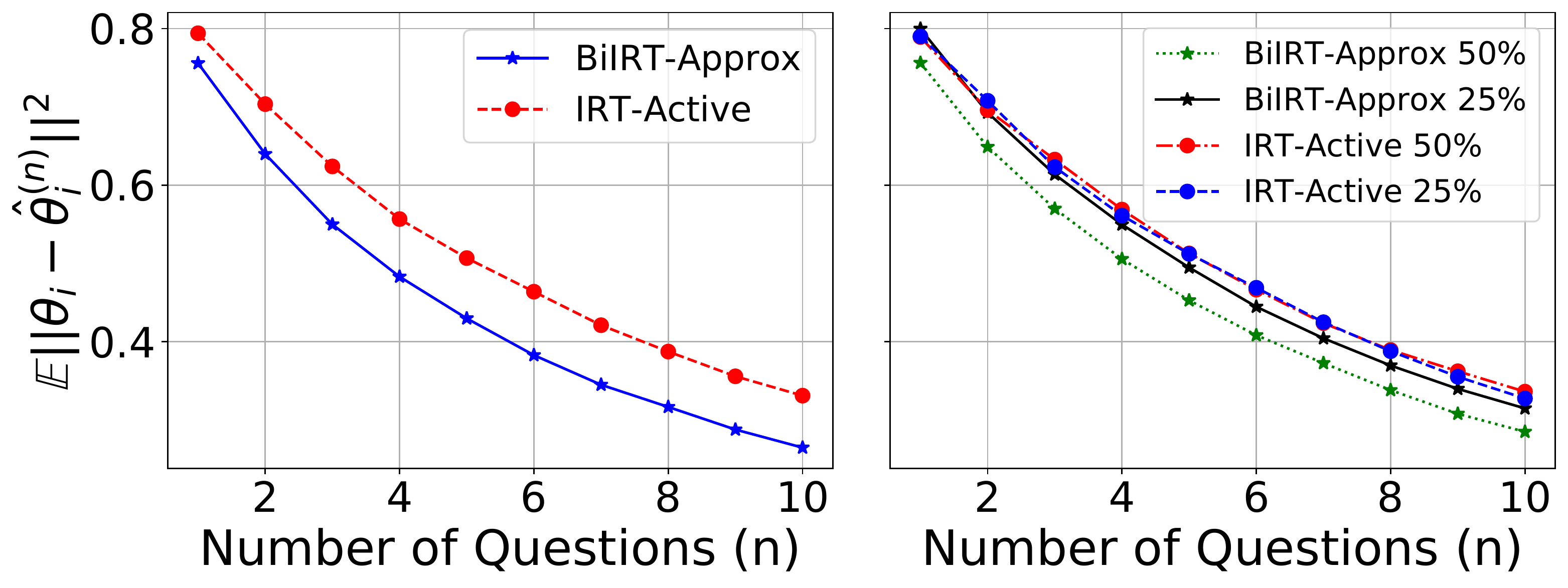}
\caption{Ability estimation accuracy on the Eedi-2 dataset.}
    \label{fig:irt-params}
\end{figure}

\paragraph{Study: Ability Estimation.}
The goal of existing real-world CAT systems is to accurately estimate the student ability parameter under IRT models, which is then used for scoring. 
Therefore, we conduct an additional experiment on the Eedi-2 dataset using the squared error between the current ability parameter estimate $\hat{\theta}_i^{(n)}$ and the true ability ${\theta}_i$ as the evaluation metric. 
Since the true student ability is unknown in real student response datasets, we use the ability value estimated from all questions each student responded to as a substitute. 
We compare two methods: IRT-Active, with the underlying 1PL IRT model trained on the data and BiIRT-Approx, where we only use the learned model-agnostic question selection algorithm for evaluation in the setting of existing CAT methods. 
Figure~\ref{fig:irt-params}(left) shows the ability estimation error (averaged over five folds) for different numbers of questions selected, $n$. 
We see that even though the goal of BiIRT-Approx is not ability parameter estimation, it is more effective than IRT-Active and can reach the same accuracy using up to $30\%$ less questions, significantly reducing test length. Figure~\ref{fig:irt-params}(right) shows the same comparison for models trained on $25\%$ and $50\%$ of the training data set. We see that BOBCAT can improve significantly as more training data becomes available while existing CAT methods cannot. 
 
\paragraph{Study: Question Exposure and Content Overlap.}
\label{sec:qualitative}
  \begin{table}[t]\centering
 \scalebox{0.75}{
     \begin{tabular}{llll}
     \toprule
Method & Exposure (median) & Exposure ($>\!20\%$) & Overlap (mean) \\\midrule
IRT-Active & 0.51\% &  0.25\% & 6.03\%\\
BiNN-Approx & 0\% & 1.54\% & 28.64\%\\ \toprule
         \end{tabular}
}
\caption{Question exposure and test overlap rates for the IRT-Active and BiNN-Approx methods on the Eedi-2 dataset. 
}
\label{tab:exposure}
\end{table}
In Table~\ref{tab:exposure}, we provide summary statistics on the question exposure rate (proportion of times a question was selected) and the test overlap rate (overlap among questions selected for two students) on the Eedi-2 dataset.  We see that BOBCAT results in a slightly higher question exposure rate than existing CAT methods but still leads to an acceptable portion of overexposed questions (more than the recommended limit of $20\%$ \cite{exposure}). However, BOBCAT results in a much higher test overlap rate than existing CAT methods \cite{stocking}. The reason for this observation is that BOBCAT favors a small subset of questions that are highly predictive of student responses to other questions, which we explain in detail next. 
Therefore, it is important for future work to develop constrained versions of BOBCAT to minimize its question exposure and test overlap rates before deployment.

 \begin{figure}[tp]
    \centering
    \includegraphics[width=.3\textwidth]{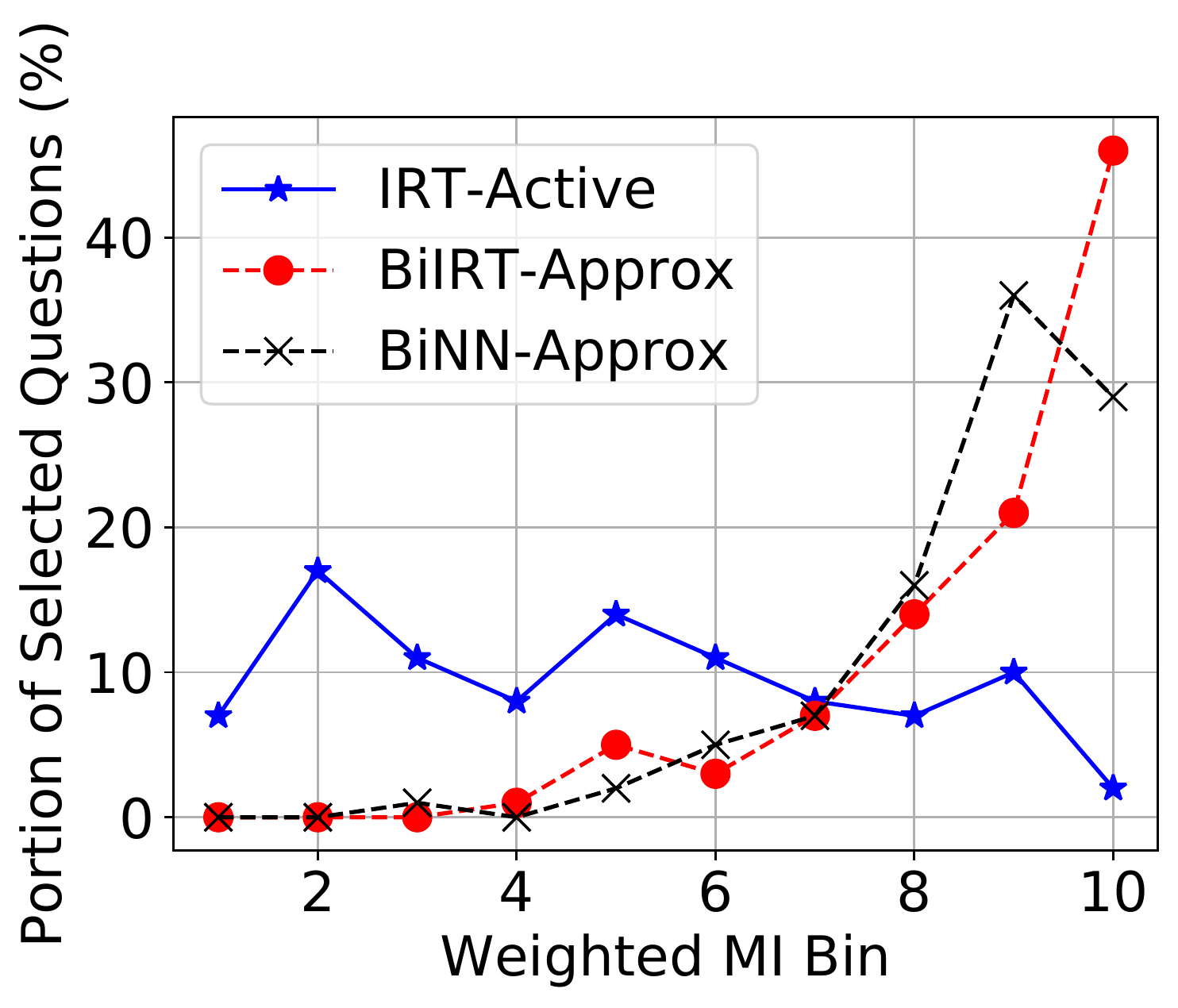}
\caption{The relationship between questions selected by each method and the mutual information between them and all other questions in the Eedi-2 dataset. BOBCAT tends to select questions that are more informative.}
    \label{fig:mi}
\end{figure}
\paragraph{Study: Question Selection.} To gain deeper insights on why BOBCAT leads to more accurate question prediction but higher question exposure and test overlap rates, we conduct a qualitative investigation. 
One possible explanation is that the gradient update in (\ref{eq:approx}) results in higher weights for questions that are more similar to the questions in the meta set (that are randomly selected out of all questions). 
Therefore, the BiIRT/NN-Approx methods favor questions that are highly representative of the entire set of questions, resulting in a higher test overlap rate in Table~\ref{tab:exposure}. 
We investigate the questions selected by the IRT-Active, BiIRT-Approx, and BiNN-Approx methods on the Eedi-2 dataset. 
We compute the weighted mutual information (MI) between each question and all others questions. 
We then use this score to assign each question to an ordered bin (from $1$ to $10$) according to their MI such that each bin has equally many questions. 
Specifically (with the following notations that apply to this study only), let $i_1,\cdots, i_m$ be the set of $m$ students who responded to both questions $j$ and $k$. 
The responses to questions $j$ and $k$ are denoted as $Y_{i_1,j},\cdots, Y_{i_m,j}$ and $Y_{i_1,k},\cdots, Y_{i_m,k}$. 
The mutual information $ MI(j,k)$ between questions $j$ and $k$ is computed as 
\begin{align*}
\sum_{x\in \{0,1\}}\!\sum_{y \in \{0,1\}}\!  \frac{|Y_{i_{\cdot},j}\!=\!x,\! Y_{i_{\cdot},k}\!=\!y |}{m}
    \log \frac{m|Y_{i_{\cdot},j}\!=\!x,\!Y_{i_{\cdot},k}\!=\!y |}{|Y_{i_{\cdot},j}\!=\!x||Y_{i_{\cdot},k}\!=\!y |}.
\end{align*}
We also compute the empirical frequency, $p_j$, of each question in the dataset, which is simply the portion of students that have responded to this question.
The weighted MI of a single question $j$ is then given by $\sum_{k, k\neq k}p_k MI(j,k)$. 
Intuitively, if the MI between a question and other questions in the dataset is high, i.e., in a higher MI bin, the question provides more information about each student than other questions.  
In Figure~\ref{fig:mi}, we plot the fraction of questions selected by the IRT-Active, BiIRT-Approx, and BiNN-Approx method from each bin for students in the test set. 
These results confirms our intuition that the BiIRT-Approx and the BiNN-Approx methods favor questions that provide more information on all other questions for every student. 
On the contrary, we do not observe such trends for the IRT-Active method; the selected questions are evenly distributed in each bin. 
These trends explain why we observe lower question exposure and test overlap rates for the IRT-Active method in Table~\ref{tab:exposure}. 
Therefore, accurately predicting student responses to questions on a long test and selecting diverse questions to minimize question exposure and test overlap rates are conflicting objectives that need to be balanced. 
This observation further highlights the need to develop constrained versions of BOBCAT that can balance these objectives. 
\section{Conclusions and Future Work}
In this paper, we proposed BOBCAT, a bilevel optimization framework for CAT, which is agnostic of the underlying student response model and learns a question selection algorithm from training data. Through extensive experiments on five real-world student response datasets, we demonstrated that BOBCAT can significantly outperform existing CAT methods at reducing test length. %
Avenues of future work include i) incorporating question exposure and content balancing  constraints \cite{content} into BOBCAT to make it deployable in real-world tests and ii) studying the fairness aspects of BOBCAT due to potential biases in training data.  
\section*{Acknowledgements}
We thank the National Science Foundation for their support under grant IIS-1917713 and Stephen Sireci for helpful discussions.
{\small
\bibliographystyle{named}
\bibliography{ijcai21}
}
\begin{center}
{\LARGE \bf 
Supplementary Material}
\end{center}

\section{Unbiased Gradient Full Objective}
 We will use the term \emph{reward} to denote the negative loss value $-\CL$.
In (\ref{eq:reinforce}), the gradient is computed based on the loss on the meta question set and the selected question probability. In practice, we can use an actor-critic network, where the actor network computes the question selection probabilities $\Pi(\cdot|\bx_i^{(t)};\bm{\phi})$ and the critic network predicts a scalar expected reward value  $V(\bx_i^{(t)})$.  The advantage value for the selected question $j_i^{(t)}$ is defined as,
\begin{align*}
    A(\bx_i^{(t)}, j_i^{(t)}) = -(\CL(\bm{\theta}_i^{\ast}, \Gamma_i)-b_i) - V(\bx_i^{(t)}).
\end{align*}
The actor network  updates the network $\bm{\phi}$ to improve the reward (or equivalently the advantage value). 
We use proximal policy optimization (PPO) for learning the unbiased model. The PPO model, a off-policy policy gradient method, takes multiple gradient steps  on the parameter $\bm{\phi}$. Since, after the first step, the updated model $\bm{\phi}$ is different than the model $\bm{\phi}_{\text{old}}$, used to compute the reward, we need to adjust the importance weights to get the unbiased gradient \cite{ppo}. 
The proximal policy optimization objective at time step $t$ for the actor network is,
\begin{align}
L_1=&-\BE_{j_i^{(t)}\sim\Pi(\bm{\phi}_{\text{old}})} \Big[\min\Big\{\frac{ \Pi(j_i^{(t)}|\bx_i^{(t)};\bm{\phi})}{\Pi (j_i^{(t)}|\bx_i^{(t)};\bm{\phi}_{\text{old}})}   A(\bx_i^{(t)}, j_i^{(t)}),\label{eq:ppo}\\
&\,\text{Clip}\big\{\frac{ \Pi (j_i^{(t)}|\bx_i^{(t)};\bm{\phi})}{\Pi (j_i^{(t)}|\bx_i^{(t)};\bm{\phi}_{\text{old}})}, 1-\epsilon, 1+\epsilon\big\}   A(\bx_i^{(t)}, j_i^{(t)})        \Big\} \Big] ,\nonumber
\end{align}
where the function $\text{Clip}(r, 1-\epsilon,1+\epsilon)$ returns $\min\{1+\epsilon, \max\{r, 1-\epsilon\}\}$; this constraint stabilizes the network by not taking large gradient steps. In addition, to encourage exploration, PPO adds an entropy objective based on the actor output probability distribution,
\begin{align}
L_2=\sum_{j\in \Omega_i^{(t)}}\Big[\Pi(j|\bx,\bm{\phi})\log \Pi(j|\bx_i^{(t)},\bm{\phi})\Big].
\end{align}
The critic network updates the parameter based on the MSE loss between the expected reward and the true reward,
\begin{align}
L_3=    ||V(\bx_i^{(t)}) +(\CL(\bm{\theta}^{\ast}, \Gamma_i)-b_i) ||^2.
\end{align}
We compute the reward for a student based on the accuracy on the meta question set. Moreover, we observed that the performance of each student differs considerably and the widely-used single moving average baseline does not perform that well; thus, we decided to use a different baseline for each student. We use meta-set performance based on a random selection algorithm as the baseline for each student. The baseline computation increases the computational complexity; however, the critic network output $V(\bx_i^{(t)})$ only needs to predict how good the actor network is compared to a random selection algorithm. This choice of baseline worked well in all of our experiments. The final objective of PPO uses a weighted combination of these individual loss terms ($L_1+0.01L_2+0.5L_3$).

 \begin{table}[t]\centering
 \scalebox{0.9}{
     \begin{tabular}{p{2cm}|p{1.4cm}|p{1.3cm}|p{1.8cm}|l}\toprule
    Dataset & Students & Questions & Interactions & R\\
    \toprule
        EdNet & 312,372 ($\sim$0.8M) & 13,169 & 76,489,425 ($\sim$95M) & 1\\\midrule
Junyi & 52224 (72,758) & 25,785 & 13,603,481 ($\sim$16M) & 2\\\midrule
Eedi-1 & 118,971 & 27,613 & 15,867,850 & 2\\ \midrule
    Eedi-2 & 4,918 & 948 & 1,382,727 & 5\\\midrule
    ASSISTments & 2,313 (4,217) & 26,688 & 325,359 (346,860) & 10\\
    \midrule
\end{tabular}
}
\caption{Full dataset details. Students with $<20$ interactions are removed. In parenthesis, we list the numbers before preprocessing/filtering. R is the repetition number of the training-meta partition split to reduce variance.}
\label{tab:dataset-extra}
\end{table}

\section{Datasets and Preprocessing Details}
We compare our models on five benchmark public datasets: EdNet, Junyi, Eedi-1, Eedi-2, and ASSISTments; the Eedi-2 dataset has been used in \cite{neuripschal}.
We remove students having less than 20 interactions; further, in case of duplicate questions, we keep only the first interaction. 
In Table~\ref{tab:dataset-extra}, we list the number of students, the number of questions, and the number of interactions. 
We also list the number of total students and interactions in each dataset in parentheses (before filtering); we do not need to preprocess Eedi-1 and Eedi-2 datasets, since they contain single interaction for each question and maintain more than $50$ total interactions for each student. 
Following \cite{neuripschal}, for smaller datasets, for each students in the validation and testing set, we created multiple  ($R$) partitions of training ($\Omega_i^{(1)}$, 80\%) and meta set ($\Gamma_i$, 20\%) to reduce the variance in the final estimate; the number of repetitions ($R$) for each dataset is added in Table~\ref{tab:dataset-extra}. {\bf The (training-meta question set) partitions for the validation and testing students are exactly the same for all models}. %

\section{Networks and Hyper-parameters}
For training IRT-Active and IRT-Random, we use $l_2$-norm regularization $\lambda_1\in\{10^{-3},10^{-6},10^{-10}\}$.  For the IRT-Active, we compute the student ability parameters using the responses and a $l_2$-norm regularization $\lambda_2\in\{10,1, 10^{-1},10^{-2},10^{-4},0\}$ to penalize deviation from the mean student ability parameter in the training dataset.
For the BiNN model, the global response model $\bm{\gamma}$ consists of a input vector $\bw\in \BR^{256}$, and a network with one hidden layer of 256 nodes with weight $\BW^1\in \BR^{256\times 256}$, and a output layer with weight $\BW^2\in \BR^{Q\times 256}$ (for simplicity, we are ignoring the biases). The response model computes probability of correctness for question $j$ with student-specific parameter $\bm{\theta}_i:=\{\bw_i, \BW_i^1, \BW_i^2\}$ as, 
    $g(j,\bm{\theta}_i) = \sigma(\BW^2_i\,\text{Dropout}(\text{ReLU}(\BW^1_i\bw_i)))[j]$,
where $\bx[j]$ represents the $j^{\text{th}}$ dimension of the vector $\bx$, and the sigmoid function $\sigma(\cdot)$ operates element-wise. We keep part of the network ($\BW^1,\BW^2$) fixed in the inner-level and only adapt the student-specific parameter $\bw$.
We use Adam optimization for all question specific parameters in BiIRT and BiMLP models with the learning rate set to  $10^{-3}$. For the rest of the parameters of $\bm{\gamma}$ (student-specific, that are adapted in the inner-level), we optimize the global values using SGD optimization with the learning rate set to $10^{-4}$ and momentum set to $0.9$. We use a fixed $K=5$ number of GD steps to optimize the inner-level objective; we find student-specific parameters quickly converge within a few steps with a slightly high learning rate $\alpha\in \{0.05, 0.1, 0.2\}$. The question selection network parameters are optimized  using Adam optimizer with learning rate $\eta_2\in \{0.002, 0.0002\}$. For the unbiased gradient, we take 4 off-policy gradient steps at the end of the selection process, and set the clipping parameter $\epsilon$ to 0.2 in all of our experiments, following \cite{ppo}. 
We set batch size fixed for all models in each dataset based on the GPU memory and train all the models until the validation performances do not improve for some number of epochs. The required GPU memory is proportional to the number of questions. We fix batch size to 128 for the Eedi-1, Junyi, and ASSISTments datasets, fix batch size to 200 for the EdNet dataset and fix batch size to 512 for the Eedi-2 dataset.

For BiIRT/BiNN-Active methods, we need to take $K=5$ GD steps in the inference time whereas -Approx and -Unbiased amortizes the cost using the neural network $\bm{\phi}$. Thus, we observe higher inference time for the -Active method compared to -Approx and -Unbiased methods.
On average, training the  -Approx and -Unbiased model takes 10-20 hours, training the BiIRT/BiNN-Active models take 15-30 hours, and training the BiIRT/BiNN-Random models take 5-10 hours  on the larger datasets (Eedi-1, Junyi, and EdNet). %

\section{Additional Experimental Results}
\begin{figure}[!htb]
    \centering
    \includegraphics[width=\columnwidth]{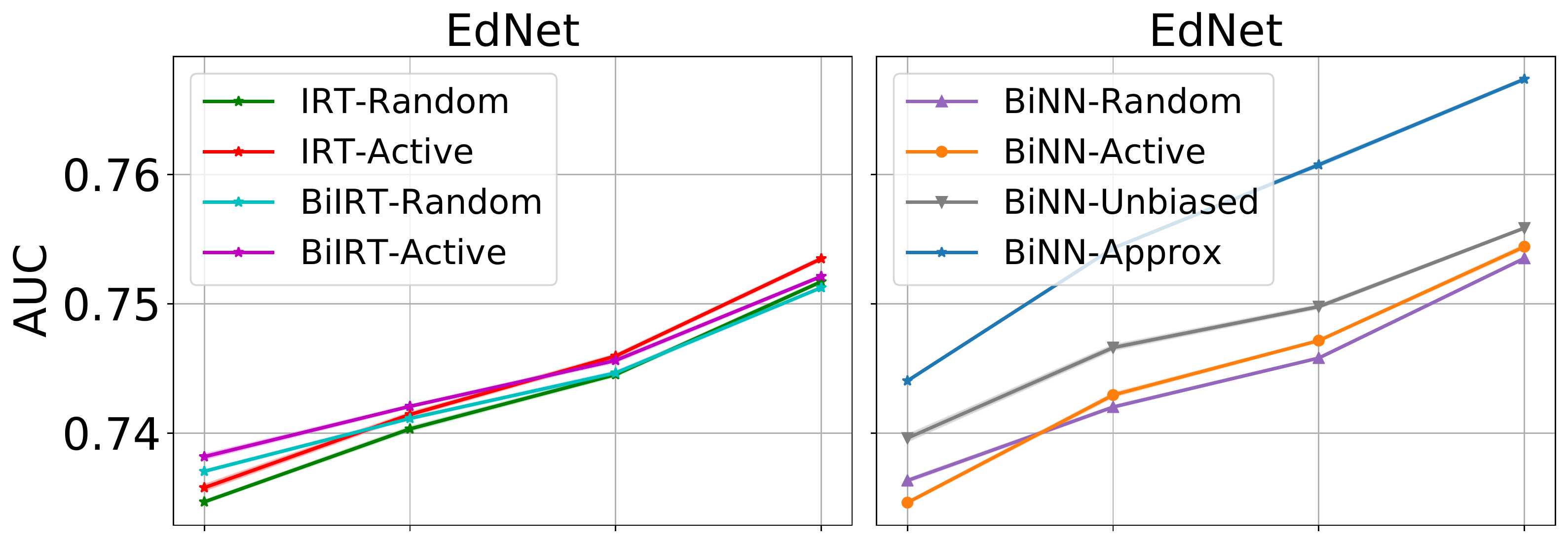}\\
    \includegraphics[width=\columnwidth]{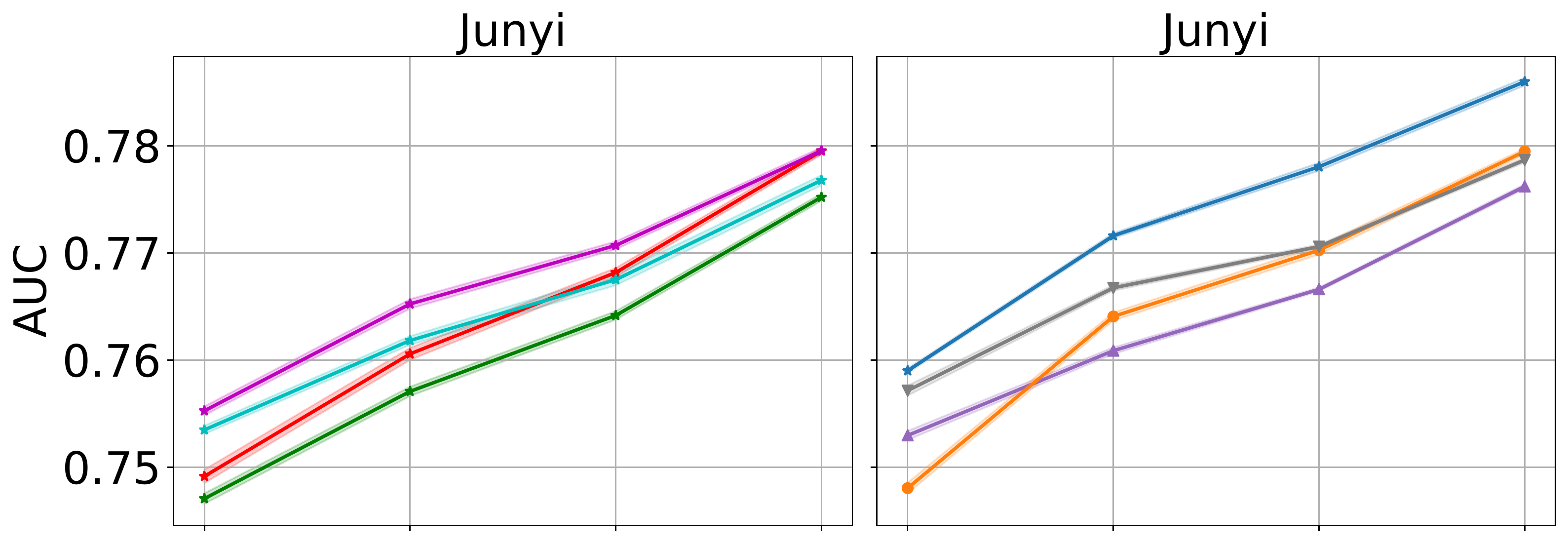}\\
    \includegraphics[width=\columnwidth]{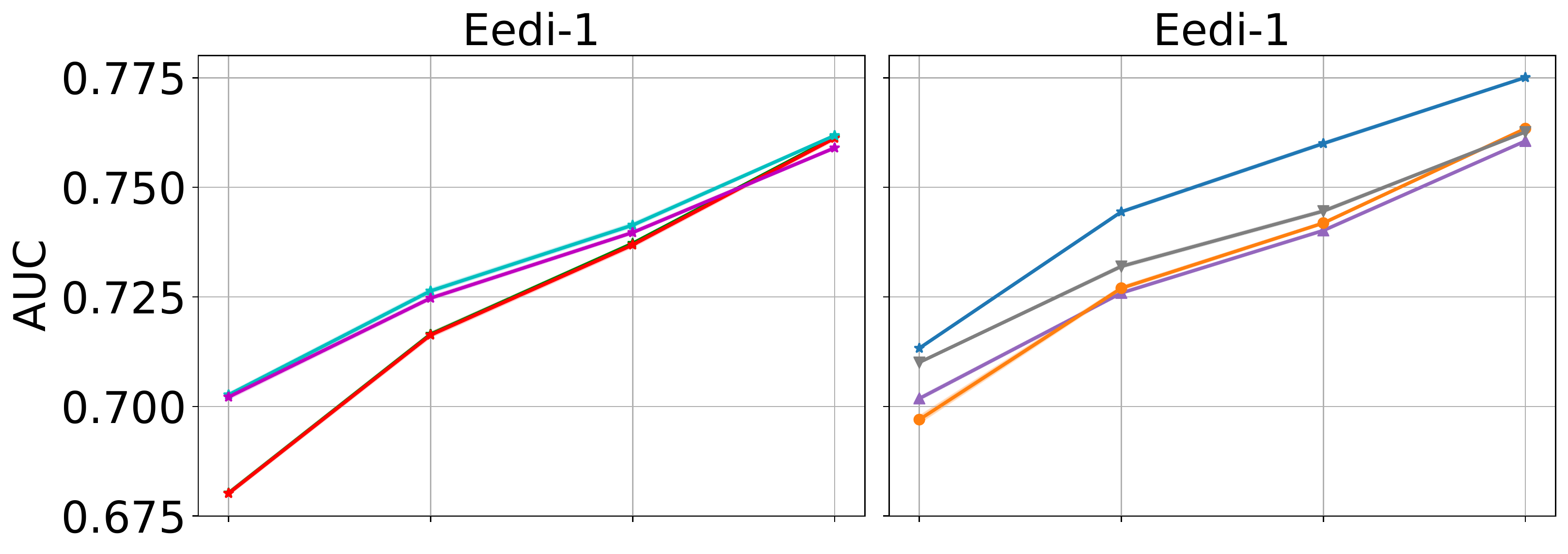}\\
    \includegraphics[width=\columnwidth]{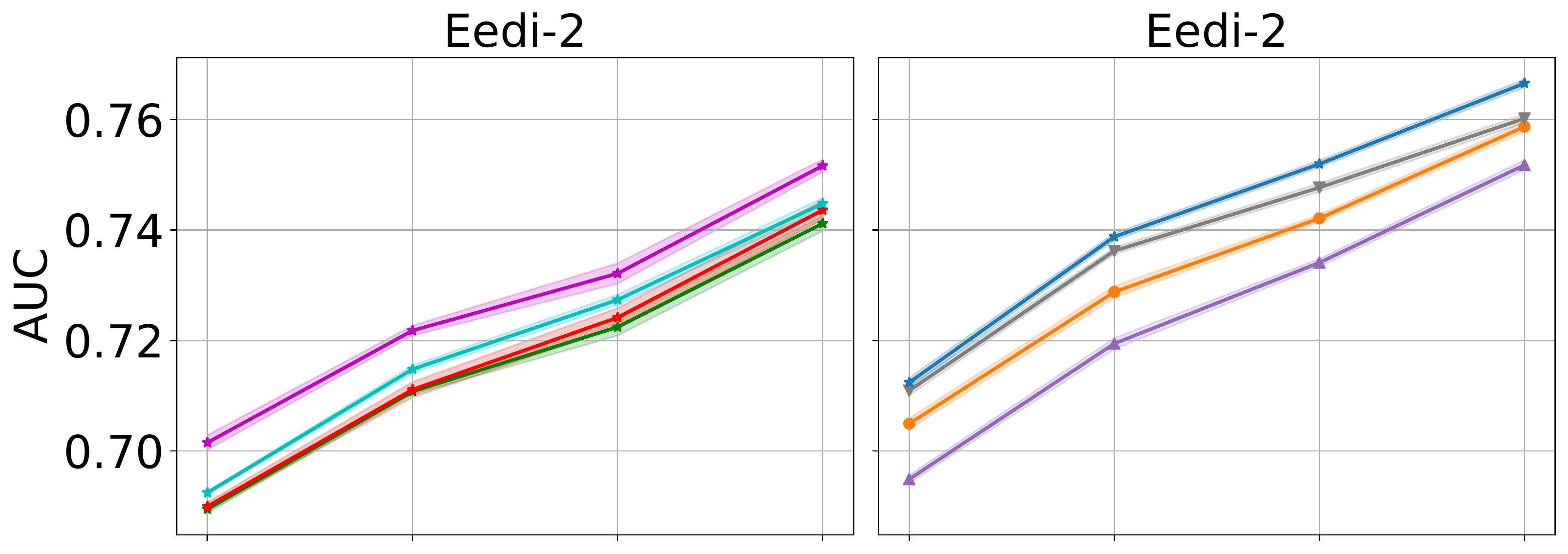}\\
    \includegraphics[width=\columnwidth]{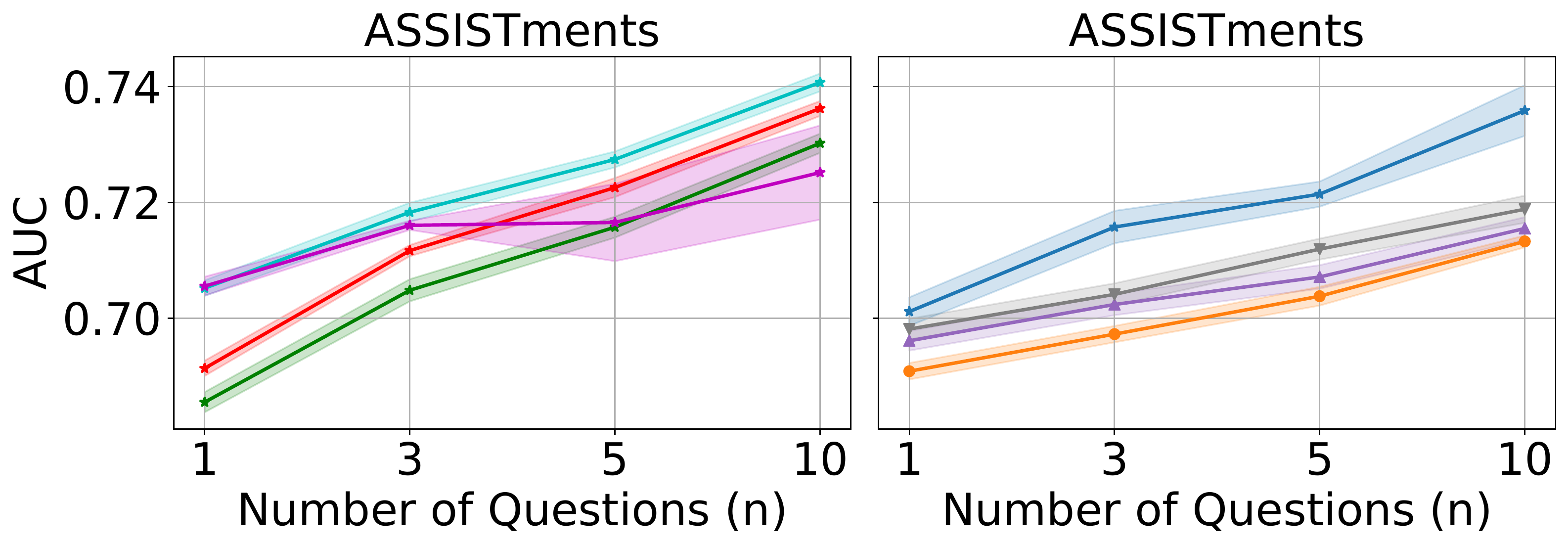}\\
    \caption{Average AUC (dark lines) and 5-fold standard deviation (light fill lines) on all datasets. First column compares IRT vs BiIRT models; second column compares all BiNN models. 
    }
    \label{fig:auc}
\end{figure}

In Table~\ref{tab:all-res-acc}, we list 5-fold mean and std accuracy numbers for all models on all datasets. Further, in Table~\ref{tab:all-res}, we list 5-fold mean and std AUC numbers for all models on all datasets. We observe similar trends for both of these metric. In Figure~\ref{fig:auc}, we provide our ablation experiment plots using AUC metric. 

 \begin{table*}[t]\centering
 \scalebox{0.7}{
     \begin{tabular}{cc|cc|cccc|cccc}\toprule
{\scriptsize Dataset} & {\scriptsize Sample} & {\scriptsize IRT-Random} & {\scriptsize IRT-Active} & {\scriptsize BiIRT-Random} & {\scriptsize BiIRT-Active} & {\scriptsize BiIRT-Unbiased} & {\scriptsize BiIRT-Approx} & {\scriptsize BiNN-Random} & {\scriptsize BiNN-Active} & {\scriptsize BiNN-Unbiased} & {\scriptsize BiNN-Approx} \\ \toprule
\multirow{4}{*}{EdNet} & 1  & 73.47$ \pm $0.01 & \bf{73.58$ \pm $0.02 }  & 73.71$ \pm $0.01 & 73.82$ \pm $0.02 & 74.14$ \pm $0.02 & \bf{74.34$ \pm $0.01 }  & 73.63$ \pm $0.01 & 73.46$ \pm $0.01 & 73.96$ \pm $0.03 & \bf{74.41$ \pm $0.01 }  \\
 & 3  & 74.03$ \pm $0.02 & \bf{74.14$ \pm $0.02 }  & 74.11$ \pm $0.01 & 74.21$ \pm $0.01 & 74.49$ \pm $0.01 & \bf{75.26$ \pm $0.01 }  & 74.2$ \pm $0.01 & 74.3$ \pm $0.02 & 74.66$ \pm $0.02 & \bf{75.43$ \pm $0.01 }  \\
 & 5  & 74.45$ \pm $0.01 & \bf{74.6$ \pm $0.02 }  & 74.47$ \pm $0.01 & 74.56$ \pm $0.02 & 74.77$ \pm $0.01 & \bf{75.68$ \pm $0.01 }  & 74.58$ \pm $0.01 & 74.72$ \pm $0.01 & 74.98$ \pm $0.02 & \bf{76.07$ \pm $0.01 }  \\
 & 10  & 75.17$ \pm $0.01 & \bf{75.35$ \pm $0.01 }  & 75.13$ \pm $0.01 & 75.21$ \pm $0.01 & 75.39$ \pm $0.01 & \bf{76.35$ \pm $0.01 }  & 75.35$ \pm $0.01 & 75.44$ \pm $0.02 & 75.59$ \pm $0.01 & \bf{76.74$ \pm $0.01 }  \\
\midrule
\multirow{4}{*}{Junyi} & 1  & 74.71$ \pm $0.05 & \bf{74.92$ \pm $0.06 }  & 75.35$ \pm $0.04 & 75.53$ \pm $0.04 & 75.67$ \pm $0.04 & \bf{75.91$ \pm $0.03 }  & 75.3$ \pm $0.04 & 74.81$ \pm $0.05 & 75.72$ \pm $0.05 & \bf{75.9$ \pm $0.02 }  \\
 & 3  & 75.71$ \pm $0.04 & \bf{76.06$ \pm $0.06 }  & 76.18$ \pm $0.04 & 76.52$ \pm $0.04 & 76.71$ \pm $0.04 & \bf{77.11$ \pm $0.04 }  & 76.09$ \pm $0.03 & 76.41$ \pm $0.04 & 76.67$ \pm $0.03 & \bf{77.16$ \pm $0.02 }  \\
 & 5  & 76.42$ \pm $0.04 & \bf{76.82$ \pm $0.04 }  & 76.75$ \pm $0.05 & 77.07$ \pm $0.04 & 77.07$ \pm $0.04 & \bf{77.69$ \pm $0.04 }  & 76.66$ \pm $0.02 & 77.03$ \pm $0.04 & 77.06$ \pm $0.02 & \bf{77.8$ \pm $0.04 }  \\
 & 10  & 77.52$ \pm $0.03 & \bf{77.95$ \pm $0.03 }  & 77.68$ \pm $0.05 & 77.95$ \pm $0.03 & 77.86$ \pm $0.04 & \bf{78.45$ \pm $0.03 }  & 77.62$ \pm $0.02 & 77.95$ \pm $0.03 & 77.87$ \pm $0.02 & \bf{78.6$ \pm $0.04 }  \\
\midrule
\multirow{4}{*}{Eedi-1} & 1  & \bf{68.03$ \pm $0.04 }  & 68.02$ \pm $0.04 & 70.27$ \pm $0.04 & 70.22$ \pm $0.05 & 70.95$ \pm $0.02 & \bf{71.34$ \pm $0.04 }  & 70.18$ \pm $0.03 & 69.7$ \pm $0.08 & 71.01$ \pm $0.02 & \bf{71.33$ \pm $0.02 }  \\
 & 3  & \bf{71.65$ \pm $0.04 }  & 71.63$ \pm $0.05 & 72.64$ \pm $0.05 & 72.47$ \pm $0.04 & 73.26$ \pm $0.06 & \bf{74.21$ \pm $0.03 }  & 72.59$ \pm $0.03 & 72.7$ \pm $0.04 & 73.2$ \pm $0.04 & \bf{74.44$ \pm $0.03 }  \\
 & 5  & \bf{73.72$ \pm $0.03 }  & 73.69$ \pm $0.05 & 74.14$ \pm $0.05 & 73.97$ \pm $0.03 & 74.54$ \pm $0.03 & \bf{75.47$ \pm $0.03 }  & 74.01$ \pm $0.03 & 74.18$ \pm $0.03 & 74.46$ \pm $0.03 & \bf{76.0$ \pm $0.03 }  \\
 & 10  & \bf{76.14$ \pm $0.02 }  & 76.12$ \pm $0.04 & 76.18$ \pm $0.02 & 75.9$ \pm $0.03 & 76.34$ \pm $0.02 & \bf{77.07$ \pm $0.03 }  & 76.06$ \pm $0.02 & 76.34$ \pm $0.02 & 76.26$ \pm $0.02 & \bf{77.51$ \pm $0.02 }  \\
\midrule
\multirow{4}{*}{Eedi-2} & 1  & 68.95$ \pm $0.07 & \bf{69.0$ \pm $0.08 }  & 69.24$ \pm $0.03 & 70.15$ \pm $0.13 & 70.64$ \pm $0.05 & \bf{70.81$ \pm $0.08 }  & 69.49$ \pm $0.07 & 70.49$ \pm $0.1 & 71.09$ \pm $0.09 & \bf{71.24$ \pm $0.09 }  \\
 & 3  & 71.08$ \pm $0.05 & \bf{71.11$ \pm $0.14 }  & 71.48$ \pm $0.07 & 72.18$ \pm $0.09 & 73.11$ \pm $0.07 & \bf{73.37$ \pm $0.08 }  & 71.94$ \pm $0.1 & 72.88$ \pm $0.11 & 73.62$ \pm $0.06 & \bf{73.88$ \pm $0.07 }  \\
 & 5  & 72.24$ \pm $0.16 & \bf{72.42$ \pm $0.17 }  & 72.74$ \pm $0.1 & 73.21$ \pm $0.18 & 74.19$ \pm $0.06 & \bf{74.55$ \pm $0.12 }  & 73.41$ \pm $0.07 & 74.21$ \pm $0.07 & 74.77$ \pm $0.08 & \bf{75.2$ \pm $0.06 }  \\
 & 10  & 74.12$ \pm $0.14 & \bf{74.36$ \pm $0.17 }  & 74.48$ \pm $0.1 & 75.17$ \pm $0.12 & 75.37$ \pm $0.13 & \bf{75.96$ \pm $0.09 }  & 75.18$ \pm $0.09 & 75.87$ \pm $0.09 & 76.02$ \pm $0.08 & \bf{76.63$ \pm $0.08 }  \\
\midrule
\multirow{4}{*}{ASSISTments} & 1  & 68.55$ \pm $0.17 & \bf{69.14$ \pm $0.13 }  & 70.52$ \pm $0.13 & 70.55$ \pm $0.16 & 71.0$ \pm $0.16 & \bf{71.33$ \pm $0.24 }  & 69.61$ \pm $0.17 & 69.09$ \pm $0.14 & 69.81$ \pm $0.18 & \bf{70.12$ \pm $0.25 }  \\
 & 3  & 70.48$ \pm $0.19 & \bf{71.17$ \pm $0.1 }  & 71.83$ \pm $0.16 & 71.6$ \pm $0.08 & 72.35$ \pm $0.19 & \bf{73.16$ \pm $0.15 }  & 70.24$ \pm $0.19 & 69.73$ \pm $0.14 & 70.41$ \pm $0.19 & \bf{71.57$ \pm $0.28 }  \\
 & 5  & 71.57$ \pm $0.18 & \bf{72.26$ \pm $0.16 }  & 72.74$ \pm $0.14 & 71.65$ \pm $0.67 & 73.1$ \pm $0.18 & \bf{73.71$ \pm $0.19 }  & 70.71$ \pm $0.2 & 70.38$ \pm $0.16 & 71.19$ \pm $0.18 & \bf{72.14$ \pm $0.22 }  \\
 & 10  & 73.02$ \pm $0.17 & \bf{73.62$ \pm $0.13 }  & 74.07$ \pm $0.15 & 72.52$ \pm $0.81 & 74.38$ \pm $0.18 & \bf{74.66$ \pm $0.22 }  & 71.55$ \pm $0.2 & 71.33$ \pm $0.1 & 71.88$ \pm $0.23 & \bf{73.59$ \pm $0.44 }  \\
\midrule
\end{tabular}
}
\caption{5-fold mean and standard deviation for all models on all datasets using AUC metric.}
\label{tab:all-res}
 \end{table*}
 
 \begin{table*}[t]\centering
 \scalebox{0.7}{
     \begin{tabular}{cc|cc|cccc|cccc}\toprule
{\scriptsize Dataset} & {\scriptsize Sample} & {\scriptsize IRT-Random} & {\scriptsize IRT-Active} & {\scriptsize BiIRT-Random} & {\scriptsize BiIRT-Active} & {\scriptsize BiIRT-Unbiased} & {\scriptsize BiIRT-Approx} & {\scriptsize BiNN-Random} & {\scriptsize BiNN-Active} & {\scriptsize BiNN-Unbiased} & {\scriptsize BiNN-Approx} \\ \toprule
\multirow{4}{*}{EdNet} & 1  & 69.99$ \pm $0.01 & \bf{70.08$ \pm $0.02 }  & 70.87$ \pm $0.01 & 70.92$ \pm $0.01 & 71.12$ \pm $0.01 & \bf{71.22$ \pm $0.01 }  & 70.81$ \pm $0.01 & 70.76$ \pm $0.01 & 71.01$ \pm $0.02 & \bf{71.22$ \pm $0.01 }  \\
 & 3  & 70.49$ \pm $0.01 & \bf{70.63$ \pm $0.01 }  & 71.07$ \pm $0.01 & 71.16$ \pm $0.0 & 71.3$ \pm $0.01 & \bf{71.72$ \pm $0.01 }  & 71.11$ \pm $0.01 & 71.23$ \pm $0.01 & 71.4$ \pm $0.02 & \bf{71.82$ \pm $0.01 }  \\
 & 5  & 70.84$ \pm $0.01 & \bf{71.03$ \pm $0.01 }  & 71.26$ \pm $0.01 & 71.37$ \pm $0.01 & 71.45$ \pm $0.01 & \bf{71.95$ \pm $0.01 }  & 71.32$ \pm $0.01 & 71.47$ \pm $0.01 & 71.56$ \pm $0.01 & \bf{72.17$ \pm $0.01 }  \\
 & 10  & 71.41$ \pm $0.01 & \bf{71.62$ \pm $0.01 }  & 71.63$ \pm $0.01 & 71.75$ \pm $0.01 & 71.79$ \pm $0.0 & \bf{72.33$ \pm $0.01 }  & 71.75$ \pm $0.01 & 71.87$ \pm $0.01 & 71.89$ \pm $0.01 & \bf{72.55$ \pm $0.01 }  \\
\midrule
\multirow{4}{*}{Junyi} & 1  & 74.38$ \pm $0.06 & \bf{74.52$ \pm $0.07 }  & 74.8$ \pm $0.06 & 74.93$ \pm $0.05 & 74.97$ \pm $0.06 & \bf{75.11$ \pm $0.05 }  & 74.78$ \pm $0.05 & 74.55$ \pm $0.06 & 75.02$ \pm $0.06 & \bf{75.1$ \pm $0.05 }  \\
 & 3  & 74.93$ \pm $0.06 & \bf{75.19$ \pm $0.07 }  & 75.24$ \pm $0.06 & 75.48$ \pm $0.07 & 75.53$ \pm $0.06 & \bf{75.76$ \pm $0.06 }  & 75.21$ \pm $0.06 & 75.4$ \pm $0.05 & 75.55$ \pm $0.04 & \bf{75.83$ \pm $0.05 }  \\
 & 5  & 75.33$ \pm $0.06 & \bf{75.64$ \pm $0.06 }  & 75.55$ \pm $0.07 & 75.79$ \pm $0.06 & 75.75$ \pm $0.05 & \bf{76.11$ \pm $0.06 }  & 75.52$ \pm $0.05 & 75.75$ \pm $0.06 & 75.74$ \pm $0.04 & \bf{76.19$ \pm $0.05 }  \\
 & 10  & 75.96$ \pm $0.06 & \bf{76.27$ \pm $0.06 }  & 76.06$ \pm $0.06 & 76.28$ \pm $0.05 & 76.19$ \pm $0.06 & \bf{76.49$ \pm $0.06 }  & 76.05$ \pm $0.05 & 76.29$ \pm $0.06 & 76.18$ \pm $0.06 & \bf{76.62$ \pm $0.06 }  \\
\midrule
\multirow{4}{*}{Eedi-1} & 1  & 66.8$ \pm $0.05 & \bf{66.92$ \pm $0.05 }  & 68.18$ \pm $0.06 & 68.22$ \pm $0.06 & 68.61$ \pm $0.04 & \bf{68.82$ \pm $0.06 }  & 68.11$ \pm $0.05 & 67.83$ \pm $0.08 & 68.66$ \pm $0.04 & \bf{68.78$ \pm $0.04 }  \\
 & 3  & 68.63$ \pm $0.05 & \bf{68.79$ \pm $0.06 }  & 69.39$ \pm $0.05 & 69.45$ \pm $0.05 & 69.81$ \pm $0.05 & \bf{70.3$ \pm $0.05 }  & 69.33$ \pm $0.05 & 69.53$ \pm $0.05 & 69.71$ \pm $0.05 & \bf{70.45$ \pm $0.05 }  \\
 & 5  & 69.94$ \pm $0.04 & \bf{70.15$ \pm $0.07 }  & 70.21$ \pm $0.05 & 70.28$ \pm $0.05 & 70.47$ \pm $0.04 & \bf{70.93$ \pm $0.05 }  & 70.12$ \pm $0.04 & 70.37$ \pm $0.05 & 70.39$ \pm $0.04 & \bf{71.37$ \pm $0.05 }  \\
 & 10  & 71.48$ \pm $0.03 & \bf{71.72$ \pm $0.05 }  & 71.44$ \pm $0.04 & 71.45$ \pm $0.04 & 71.57$ \pm $0.05 & \bf{72.0$ \pm $0.05 }  & 71.32$ \pm $0.04 & 71.7$ \pm $0.05 & 71.44$ \pm $0.03 & \bf{72.33$ \pm $0.04 }  \\
\midrule
\multirow{4}{*}{Eedi-2} & 1  & 63.65$ \pm $0.12 & \bf{63.75$ \pm $0.12 }  & 64.16$ \pm $0.07 & 64.83$ \pm $0.12 & 65.22$ \pm $0.04 & \bf{65.3$ \pm $0.08 }  & 64.31$ \pm $0.07 & 65.1$ \pm $0.09 & 65.55$ \pm $0.07 & \bf{65.65$ \pm $0.06 }  \\
 & 3  & 65.24$ \pm $0.11 & \bf{65.25$ \pm $0.19 }  & 65.84$ \pm $0.09 & 66.42$ \pm $0.09 & 67.09$ \pm $0.09 & \bf{67.23$ \pm $0.08 }  & 66.15$ \pm $0.1 & 66.94$ \pm $0.1 & 67.55$ \pm $0.06 & \bf{67.79$ \pm $0.05 }  \\
 & 5  & 66.24$ \pm $0.12 & \bf{66.41$ \pm $0.13 }  & 66.72$ \pm $0.1 & 67.35$ \pm $0.1 & 67.91$ \pm $0.07 & \bf{68.23$ \pm $0.09 }  & 67.29$ \pm $0.06 & 67.98$ \pm $0.06 & 68.43$ \pm $0.07 & \bf{68.82$ \pm $0.05 }  \\
 & 10  & 67.71$ \pm $0.1 & \bf{68.04$ \pm $0.12 }  & 68.04$ \pm $0.1 & 68.99$ \pm $0.09 & 68.84$ \pm $0.1 & \bf{69.47$ \pm $0.06 }  & 68.68$ \pm $0.07 & 69.34$ \pm $0.08 & 69.47$ \pm $0.05 & \bf{70.04$ \pm $0.04 }  \\
\midrule
\multirow{4}{*}{ASSISTments} & 1  & 65.45$ \pm $0.2 & \bf{66.19$ \pm $0.19 }  & 68.63$ \pm $0.24 & 68.69$ \pm $0.27 & 69.03$ \pm $0.24 & \bf{69.17$ \pm $0.3 }  & 67.89$ \pm $0.23 & 67.44$ \pm $0.18 & 67.98$ \pm $0.25 & \bf{68.0$ \pm $0.25 }  \\
 & 3  & 67.52$ \pm $0.25 & \bf{68.75$ \pm $0.15 }  & 69.42$ \pm $0.24 & 69.54$ \pm $0.21 & 69.78$ \pm $0.23 & \bf{70.21$ \pm $0.19 }  & 68.23$ \pm $0.24 & 68.13$ \pm $0.18 & 68.33$ \pm $0.22 & \bf{68.73$ \pm $0.25 }  \\
 & 5  & 68.55$ \pm $0.25 & \bf{69.87$ \pm $0.21 }  & 70.03$ \pm $0.21 & 69.79$ \pm $0.32 & 70.3$ \pm $0.26 & \bf{70.41$ \pm $0.23 }  & 68.52$ \pm $0.24 & 68.68$ \pm $0.19 & 68.7$ \pm $0.21 & \bf{69.03$ \pm $0.25 }  \\
 & 10  & 69.95$ \pm $0.2 & \bf{71.04$ \pm $0.22 }  & 70.88$ \pm $0.2 & 70.66$ \pm $0.35 & \bf{71.17$ \pm $0.21 }  & 71.14$ \pm $0.24 & 68.9$ \pm $0.21 & 69.24$ \pm $0.19 & 69.04$ \pm $0.24 & \bf{69.75$ \pm $0.3 }  \\
\midrule
\end{tabular}
}
\caption{5-fold mean and standard deviation for all models on all datasets using accuracy metric.
}
\label{tab:all-res-acc}
 \end{table*}

\end{document}